\documentclass[12pt,a4paper]{article}

\usepackage[T1]{fontenc}
\usepackage[utf8]{inputenc}
\usepackage{times}

\usepackage[margin=2.5cm]{geometry}

\usepackage{amsmath,amssymb}

\usepackage{graphicx}
\usepackage{tikz}
\usepackage{pgfplots}
\pgfplotsset{compat=1.18}
\usetikzlibrary{arrows.meta, shapes.geometric, positioning, decorations.pathmorphing,
                decorations.markings, calc, fit, backgrounds}

\usepackage{booktabs}
\usepackage{array}
\usepackage{tabularx}
\usepackage{longtable}

\usepackage{enumitem}

\usepackage{titlesec}
\titleformat{\section}{\large\bfseries}{\thesection.}{0.5em}{}
\titleformat{\subsection}{\normalsize\bfseries}{\thesubsection.}{0.5em}{}

\usepackage{xcolor}
\definecolor{darkblue}{RGB}{0,51,102}
\definecolor{medgray}{RGB}{120,120,120}
\definecolor{lightgray}{RGB}{240,240,240}
\definecolor{accentred}{RGB}{180,30,30}
\definecolor{accentgreen}{RGB}{30,130,70}
\definecolor{stageBlue}{RGB}{40,92,143}
\definecolor{stageTeal}{RGB}{44,137,151}
\definecolor{stageGreen}{RGB}{75,140,92}
\definecolor{stageAmber}{RGB}{184,128,48}
\definecolor{stagePlum}{RGB}{123,87,147}
\definecolor{stageRed}{RGB}{168,77,92}
\definecolor{flowGray}{RGB}{90,90,90}

\usepackage[colorlinks=true, linkcolor=darkblue, citecolor=darkblue,
            urlcolor=darkblue, bookmarks=true, hyperfootnotes=false]{hyperref}
\hypersetup{%
  pdfauthor={Vu Hung Nguyen; Thanh Nguyen},
  pdftitle={SDAD: Spec-Driven Agentic Development for the AI-Native SDLC}%
}

\usepackage[numbers,sort&compress]{natbib}
\usepackage{abstract}

\usepackage{fancyhdr}
\begin{document}

\newgeometry{top=1.5cm,bottom=2cm,left=2.5cm,right=2.5cm}
\begin{titlepage}
  \thispagestyle{empty}
  \centering
  \vspace*{0.6cm}
  {\LARGE\bfseries SDAD: Spec-Driven Agentic Development\\[0.2em]
   for the AI-Native SDLC\par}
  \vspace{0.85cm}
  \makeatletter
  {\large Vu Hung Nguyen\thanks{Australian Catholic University, \href{mailto:vunguyen@acu.edu.au}{vunguyen@acu.edu.au}}\par}
  {\large Thanh Nguyen\thanks{International College of Management Sydney (ICMS), \href{mailto:thnguyen@icms.edu.au}{thnguyen@icms.edu.au}}\par}
  \@thanks
  \makeatother
  \vspace{0.7cm}
  {\large\textit{Technical Report --- SDAD-Framework}\par}
  \vspace{0.35cm}
  {\normalsize May 2026\footnotemark\par}
  \footnotetext{2026 The Authors. Licensed under CC BY-SA 4.0.}
  \vspace{0.55cm}
  \par
  \begin{flushleft}
    \begingroup
    \linespread{0.94}\selectfont
    \begin{abstract}
Frontier coding agents backed by large language models with hundred-thousand- to
million-token contexts are restructuring the Software Development Life Cycle
({SDLC}). Rich context and multi-step reasoning now allow substantial
Functional Requirement Documents ({FRDs}) and repository context to be ingested
in one workflow, so specification quality returns as \emph{execution fuel} for
autonomous delivery. This report formalises \emph{Spec-Driven Agentic
Development} ({SDAD}) as a synthesis of disciplined up-front formalisation and
high-velocity implementation: intent capture, machine-readable specification,
agentic synthesis, and independent multi-agent verification under human
sign-off. We revisit the historical pendulum between Waterfall and Agile,
introduce \emph{AI-code} as a fourth production paradigm, and compare
{Human-Agile} ({circa}~2020) with {Agentic-SDAD} ({circa}~2026) across
artefacts, cadence, accountability, and security posture.

Beyond process description, we extend the model to team role metamorphosis
(engineer, QA, platform, and product functions), quantitative governance
(Ambiguity Tax, {Spec Fidelity}, {SER}, and
$\mathrm{TCI}_{\mathrm{agentic}}$ with repair multiplier~$\varphi$), and
pragmatic adoption via hybrid estimation and a staged migration blueprint.
Industrial and research evidence on {AI}-augmented testing and verification is
integrated to motivate separation between synthesis and release authority.
Overall, the paper argues that agentic speed does not eliminate engineering
discipline; it relocates discipline upstream into specification precision,
explicit gates, and auditable provenance.
    \end{abstract}
    \endgroup
  \end{flushleft}
\end{titlepage}
\restoregeometry

\tableofcontents
\newpage

\section{Introduction}

\subsection{The Paradigm Shift: From Human-Centric to Agent-Mediated
  Development}

For two decades, the dominant industry trend was toward smaller, more
iterative releases designed to mitigate the high cost of human communication
and the rigidity of the Waterfall model. The Agile Manifesto of 2001
crystallised this preference, championing ``individuals and interactions over
processes and tools.'' However, in 2026, the bottleneck is no longer the speed
of typing code, but the \textbf{clarity of intent}. With agents capable of
ingesting entire repositories and documentation sets in a single inference
pass~\cite{geminiteam2024gemini15}, the methodology must adapt to a world
where ``\emph{The Spec is the Code}.''

\subsection{The Co-evolution of Methodology and Model Scale}

This paper argues that the emergence of state-of-the-art large language models
({SOTA LLMs}---frontier systems likely exceeding 500B parameters once routing
and multimodal stacks are included, even where vendors omit published counts)
operating over hundred-thousand- to million-token context windows has
catalysed a co-evolutionary pressure on
software methodology: as model capability scales, optimal human workflow
converges toward the formal, pre-emptive specification paradigm historically
associated with Waterfall and the Rational Unified Process (RUP). We term this
convergence \emph{Spec-Driven Agentic Development} (SDAD) and argue that it
represents not a regression to bureaucratic rigidity, but a \emph{maturation}
of engineering practice in which human cognition is redirected from
implementation detail to intent management.

The remainder of this paper is structured as follows. Section~\ref{sec:history}
reviews the historical pendulum between rigidity and flexibility in software
methodology. Section~\ref{sec:waterfall} summarises the classical Waterfall
phases and how AI can shorten each phase while preserving explicit inputs,
outputs, and phase-exit criteria. Section~\ref{sec:technical} analyses the technical underpinnings
of {SOTA LLMs} and long-context engineering. Section~\ref{sec:ai-code} introduces
\emph{AI-code} as a fourth paradigm of software production and defines the
accountability question that motivates the {SDAD} governance model.
Section~\ref{sec:bduf} discusses the ``Waterfall Resurgence'' and Big Design Up
Front (BDUF) 2.0.
Section~\ref{sec:agile} examines the metamorphosis of Agile. Section~\ref{sec:framework}
presents the SDAD framework (including automated testing under agentic pipelines;
Subsection~\ref{subsec:ai-testing}). Section~\ref{sec:comparison} provides a
comparative analysis of SDLC 2020 versus SDLC 2026. Section~\ref{sec:economics}
summarises the quantitative economics of agentic synthesis versus traditional
delivery. Section~\ref{sec:discussion}
addresses ethical and organisational challenges. Section~\ref{sec:conclusion}
concludes the paper.

\section{Historical Review: The Pendulum of Rigidity and Flexibility}
\label{sec:history}

\begin{figure}[htbp]
  \centering
  \begin{tikzpicture}[scale=1.2, every node/.style={font=\small}]
    \filldraw[darkblue] (0,0) circle (3pt);

    \draw[thick, dotted, medgray]
      (-3,-3) arc[start angle=225, end angle=315, radius=4.24];

    \draw[very thick, darkblue, -{Stealth[length=7pt]}]
      (0,0) -- (-2.5,-2.5)
      node[below left, align=center]{%
        \textbf{Formal Rigidity}\\
        Waterfall / RUP\\
        (1970s--1990s)};

    \draw[very thick, darkblue, -{Stealth[length=7pt]}]
      (0,0) -- (2.5,-2.5)
      node[below right, align=center]{%
        \textbf{Iterative Flexibility}\\
        Agile / Scrum\\
        (2001--2020)};

    \filldraw[medgray] (1.3,-1.9) circle (2pt)
      node[right=4pt, medgray, align=left]{Copilot Era\\(2021--2024)};

    \draw[ultra thick, accentred, -{Stealth[length=7pt]}]
      (0,0) -- (0,-3.5)
      node[below, align=center,
           draw=accentred, rounded corners=2pt,
           inner sep=4pt, fill=white]{%
        \textbf{Automated Rigidity (SDAD)}\\
        Spec-Driven Synthesis\\
        (2025--2026)};

    \draw[<->, thin, medgray] (-3.2,-5.6) -- (3.2,-5.6);
    \node[medgray, anchor=north west] at (-3.2,-5.65) {\scriptsize High Fidelity};
    \node[medgray, anchor=north east] at ( 3.2,-5.65) {\scriptsize High Velocity};
    \node[medgray, anchor=north]      at (0,-5.65)    {\scriptsize Documentation~$\longleftrightarrow$~Velocity};
  \end{tikzpicture}
  \caption{The methodological pendulum swings from Formal Rigidity
    (Waterfall/RUP, 1970s--1990s)~\cite{royce1970} through Iterative Flexibility
    (Agile/Scrum, 2001--2020) to SDAD as a synthesis that achieves high
    documentation fidelity \emph{and} high execution velocity simultaneously.}
  \label{fig:pendulum}
\end{figure}
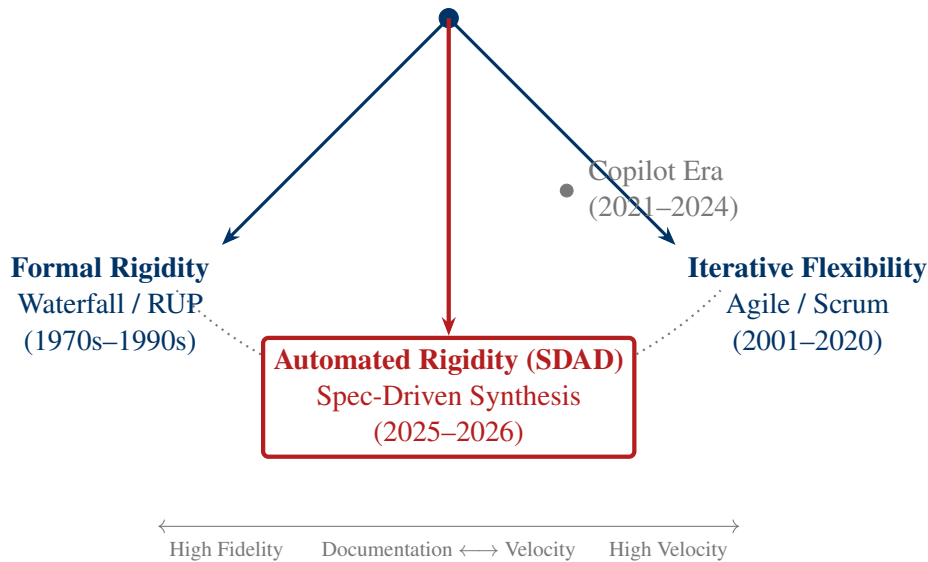

The history of software methodology can be read as a pendulum swinging between
two poles: \emph{rigidity} (comprehensive up-front planning, formal
specification, and staged execution) and \emph{flexibility} (iterative
delivery, lightweight documentation, and continuous customer feedback).
Royce~\cite{royce1970} crystallised the rigidity pole for large programmes,
while Abbas \emph{et al.}~\cite{abbas2008} show that ``Agile thinking''---often
framed only as a 2000s reaction to tradition---recycles much older iterative and
critical practice from software engineering history.
Figure~\ref{fig:pendulum} visualises this historical arc.

\subsection{1970s--1990s: The Era of Determinism (Waterfall/RUP)}

The necessity of disciplined structure in low-concurrency development
environments motivated staged lifecycle thinking and its textbook ``Waterfall''
caricature. Royce~\cite{royce1970}, reflecting spacecraft mission-planning
experience, warned that analysis-and-coding alone is ``doomed to failure'' for
large systems and advocated an enlarged sequence---requirements, design,
implementation, testing---with \emph{feedback between stages}, preliminary design,
heavy documentation, prototyping, and early customer involvement. Later teaching
often compressed his diagram into a purely linear cascade; that simplification
underpins modern critiques of rigidity even though Royce's own prescription was
more iterative than the folklore suggests.

Together with successors such as the Rational Unified Process, these paradigms
treated comprehensive documentation---requirements specifications, architecture
documents, and detailed design artefacts---as primary risk-management artefacts.
\textbf{Strengths} include traceability, reviewability, and contracts that survive
staff turnover. \textbf{Weaknesses} include slow feedback when markets shift
during long freezes and cost sensitivity to late defect discovery. Waterfall's
insistence on formalised specifications nonetheless yields artefacts that behave
like machine-interpretable intent bundles---a property whose value grows sharply
once autonomous agents, rather than typing-bound humans, consume those bundles.

\subsection{2000s--2020: The Agile Manifesto \& The Rise of Scrum}

The Agile Manifesto of 2001 formalised what Abbas \emph{et al.}~\cite{abbas2008}
describe as both a reaction to heavyweight process and a recombination of older
ideas---incremental delivery, responsiveness to change, tight customer
collaboration, and scepticism toward documentation-for-its-own-sake---that had
surfaced in critiques of traditional methods long before ``Agile'' was a brand.

Operationally, Agile responded to Waterfall's ``latency of feedback'': by the
time a large specification became working code, requirements often stale-dated.
Scrum's two-week sprint compressed the loop from months to weeks. \textbf{Pros}
include faster validation with users, risk reduction via incremental releases,
and organisational transparency through inspect--adapt ceremonies.
\textbf{Cons}, intentionally accepted under human-centric throughput, include
thin or tacit specification where knowledge lives in conversations and tribal
memory--- workable when engineers hold the whole model in mind, but brittle when
machine agents must reproduce behaviour without access to undocumented intent.

\subsection{Royce, Agile, and where {AI} pushes the pendulum}

Under classical Agile, ``Working software over comprehensive documentation''
made sense because humans translated fuzzy intent into code slowly anyway.
Frontier models invert part of that calculus: they synthesise implementation at
speed that rewards \emph{durable, inspectable specification}---pulling practice
back toward Royce-style artefacts without necessarily restoring multi-year
frozen cascades~\cite{royce1970,abbas2008,he2026speed}. The synthesis pursued in
later sections is thus neither nostalgia for bureaucracy nor denial of Agile's
feedback virtues; it is a reconciliation tuned to agents as consumers of intent.

\subsection{2021--2024: The Copilot Era}

GitHub Copilot's public launch in 2021 marked the first mainstream integration
of large language models into the development workflow. In this era, AI
functioned primarily as an autocomplete utility---accelerating token-level
coding while leaving architectural and specification decisions entirely to human
engineers. The persistence of traditional Agile frameworks during this period
reflects the limited autonomous capability of Copilot-generation models:
context windows of 8,000--100,000 tokens were insufficient to ingest project-
level context, and multi-file reasoning remained unreliable~\cite{salem2024survey}.

\subsection{2025--2026: The Agentic Turn}

The period 2025--2026 marks a qualitative discontinuity. Models operating over
millions of tokens~\cite{geminiteam2024gemini15}, combined with multi-agent
orchestration frameworks, enable fully autonomous implementation of complex,
cross-module features from a single specification document. Empirical
evidence~\cite{he2026speed} documents 3--5$\times$ velocity gains for
AI-assisted teams but also identifies a critical dependency: the quality of
agentic output degrades sharply in the absence of rigorous up-front
specification. This finding is the empirical foundation of the SDAD framework.

\section{Classical Waterfall: Phases and AI-Augmented Acceleration}
\label{sec:waterfall}

\begin{figure}[htbp]
  \centering
  \resizebox{0.95\textwidth}{!}{
%

\begin{tikzpicture}[
  font=\scriptsize,
  tradbox/.style={
    rectangle,
    rounded corners=3pt,
    draw=darkblue,
    thick,
    fill=darkblue!8,
    align=center,
    inner sep=3.5pt,
    text width=4.35cm,
    minimum height=7.5mm,
  },
  aiphase/.style={
    rectangle,
    rounded corners=3pt,
    draw=darkblue!58!accentgreen,
    thick,
    fill=white,
    align=center,
    inner sep=3.5pt,
    text width=3.08cm,
    minimum height=7.5mm,
  },
  aihint/.style={
    rectangle,
    rounded corners=2pt,
    draw=accentgreen!75!black,
    fill=accentgreen!14,
    align=left,
    inner sep=2.8pt,
    font=\fontsize{7.05}{8.4}\selectfont,
    text width=4.45cm,
    minimum height=14mm,
  },
  coltitle/.style={
    font=\footnotesize\bfseries,
    anchor=south,
    text=darkblue,
    inner sep=0pt,
    align=center,
  },
  flow/.style={
    darkblue,
    very thick,
    -{Stealth[length=6pt,width=5pt]},
  },
  lite/.style={
    medgray,
    thick,
    densely dashed,
    -{Stealth[length=5pt,width=4pt]},
  },
]

  \node[coltitle, text width=4.65cm] (hdL)
    at (-6.32, -0.15) {(a)~Traditional cascade};
  \node[coltitle, text width=8.65cm] (hdR)
    at (5.18, -0.15) {(b)~AI-accelerated (same sequence)};

  \node[tradbox, anchor=north] (t1) at ($(hdL.south)+(0,-11mm)$)
    {\textbf{System requirements}};
  \node[tradbox, below=8.5mm of t1] (t2) {\textbf{Software requirements}};
  \node[tradbox, below=8.5mm of t2] (t3) {\textbf{Analysis}};
  \node[tradbox, below=8.5mm of t3] (t4) {\textbf{Program design}};
  \node[tradbox, below=8.5mm of t4] (t5) {\textbf{Coding}};
  \node[tradbox, below=8.5mm of t5] (t6) {\textbf{Testing}};
  \node[tradbox, below=8.5mm of t6] (t7) {\textbf{Operations}};

  \foreach \Fa/\Fb in {t1/t2,t2/t3,t3/t4,t4/t5,t5/t6,t6/t7}
    {\draw[flow] (\Fa.south) -- (\Fb.north);}

  \node[aiphase, anchor=north] (a1) at ($(hdR.south -| 2.52, 0)+(0,-11mm)$)
    {\textbf{System requirements}};
  \node[aiphase, below=8.5mm of a1] (a2) {\textbf{Software requirements}};
  \node[aiphase, below=8.5mm of a2] (a3) {\textbf{Analysis}};
  \node[aiphase, below=8.5mm of a3] (a4) {\textbf{Program design}};
  \node[aiphase, below=8.5mm of a4] (a5) {\textbf{Coding}};
  \node[aiphase, below=8.5mm of a5] (a6) {\textbf{Testing}};
  \node[aiphase, below=8.5mm of a6] (a7) {\textbf{Operations}};

  \foreach \Fa/\Fb in {a1/a2,a2/a3,a3/a4,a4/a5,a5/a6,a6/a7}
    {\draw[flow] (\Fa.south) -- (\Fb.north);}

  \node[aihint, anchor=mid west]
    at ($(a1.east)+(0.14,0)$) {%
      \textit{Assist:} summarise sources; glossary; stakeholder Q\,\&\,A transcripts.%
    };
  \node[aihint, anchor=mid west]
    at ($(a2.east)+(0.14,0)$) {%
      \textit{Assist:} draft SRS clauses; ambiguity or conflict reports; checklist vs.\ exit criteria.%
    };
  \node[aihint, anchor=mid west]
    at ($(a3.east)+(0.14,0)$) {%
      \textit{Assist:} trade-off options; feasibility notes; trace sketches to SRS ids.%
    };
  \node[aihint, anchor=mid west]
    at ($(a4.east)+(0.14,0)$) {%
      \textit{Assist:} architectures, diagrams, APIs; rationales mined from SRS.%
    };
  \node[aihint, anchor=mid west]
    at ($(a5.east)+(0.14,0)$) {%
      \textit{Assist:} code synthesis/refactor; policy lint; scaffolding and reviews.%
    };
  \node[aihint, anchor=mid west]
    at ($(a6.east)+(0.14,0)$) {%
      \textit{Assist:} test cases \& data; regression packs; artefact-aligned oracles.%
    };
  \node[aihint, anchor=mid west] (h7)
    at ($(a7.east)+(0.14,0)$) {%
      \textit{Assist:} runbooks \& cutover drafts; incident digests from logs \& tickets.%
    };


  \path let
    \p1 = (t7.south),
    \p2 = (h7.south),
    \n{m}={min(\y1,\y2)}
  in coordinate (figBottom) at (0,\n{m});

  \node[
    rectangle,
    draw=medgray,
    densely dashed,
    rounded corners=2pt,
    inner sep=5pt,
    font=\fontsize{8pt}{9}\selectfont,
    align=left,
    anchor=north,
    text width=15.05cm,
  ]
    at ($(figBottom)+(0,-1.06)$)
    {\textbf{Key contrast:}\ both panels preserve \emph{one-way} sequencing between terraces
      (Royce's formulation); frontier agents shorten \emph{elapsed time inside each terrace}
      through faster drafting, review, and conformance checks---while phase interfaces and
      sign-off gates remain explicit.};

\end{tikzpicture}}
  \caption{Royce's seven-phase Waterfall shown as a traditional cascade (left) and with
    \emph{per-terrace} frontier-model assistance (right). AI acceleration compresses elapsed
    effort \emph{inside} each phase; sequential sign-off across interfaces remains unchanged.}
  \label{fig:waterfall-ai}
\end{figure}
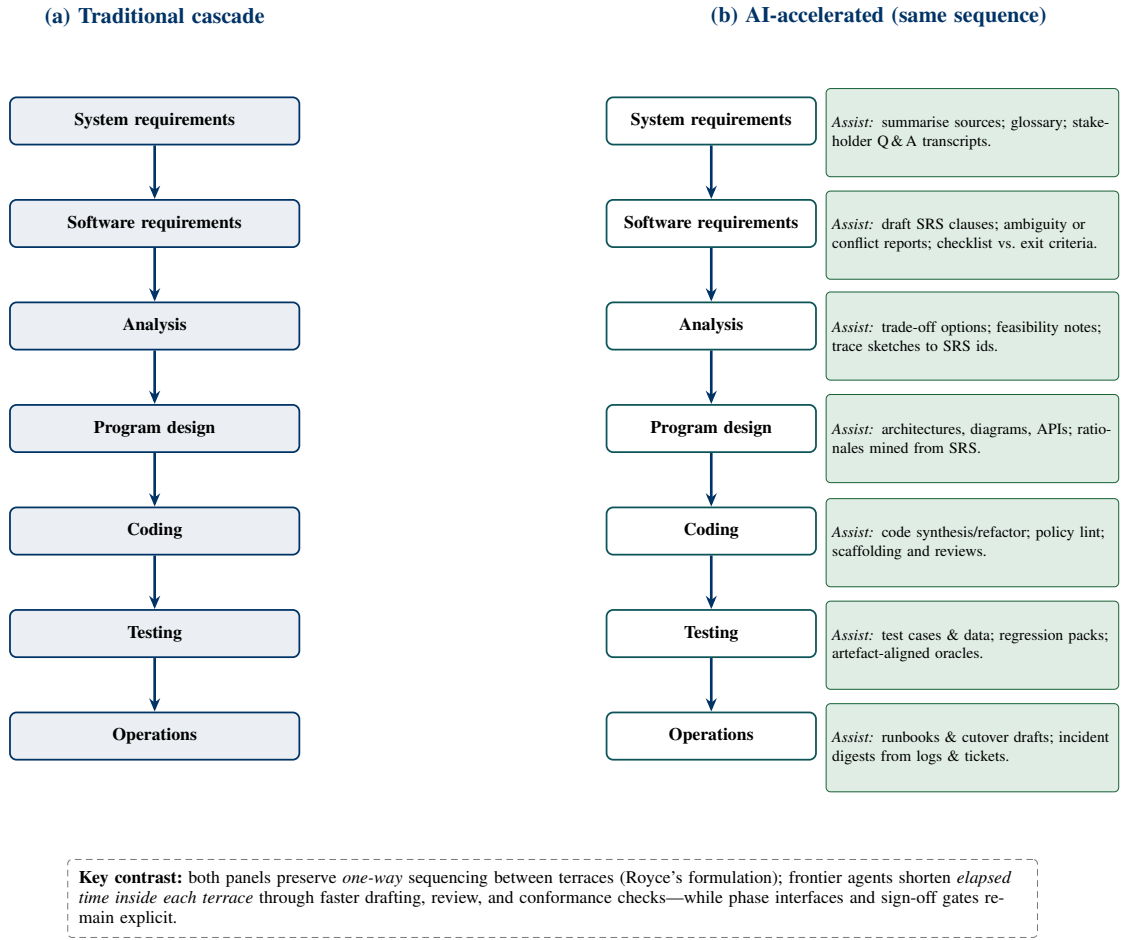

The classical Waterfall model treats software development as a largely
\textbf{sequential} progression through Royce-style terraces (\emph{e.g.}
system/software requirements through analysis to program design, coding,
testing, and operations), each consuming signed-off artefacts from the previous
terrace and yielding a named deliverable~\cite{royce1970}.
Figure~\ref{fig:waterfall-ai} compares the textbook cascade against the same ordering
\textbf{paired} with representative \textbf{agent} assistance inside every terrace:
the topological structure is unchanged, but frontier models reduce the latency and labour
of producing and checking documentation, designs, tests, and operational feedback---provided
\textbf{interfaces} (inputs and outputs) and \textbf{success criteria} (exit gates)
remain explicit~\cite{he2026speed,otoum2026methods}.

\subsection{Phases, Artefacts, and Typical AI Assistance}

\textbf{Requirements analysis and specification.}
Stakeholder intent is clarified, inconsistencies removed, and the result fixed in
an SRS (software requirements specification): the authoritative contract between
customer and builders. Agents can summarise workshops, propose requirement
structures, draft and diff SRS sections, surface ambiguities, and suggest verifiable
acceptance criteria---still subject to human sign-off before the SRS ``freezes''
the downward flow.

\textbf{Design.}
SRS content is refined into architectures and module-level behaviours recorded in
high-level design (HLD), low-level design (LLD), and the combined software
design document (SDD). Agents can synthesise diagrams, interfaces, API contracts,
runbooks-for-build, and rationale text at speed; HITL reviewers remain accountable
for trade-offs that are not mechanically derivable from the SRS alone.

\textbf{Development.}
Implementers map the SDD into source code using agreed languages and tools, with
\textbf{unit testing} validating individual modules before integration. Coding
assistants reduce implementation time sharply; autonomous multi-file pipelines
(from formal specs) amplify that effect~\cite{pohle2026agentictyper}. Human or
verification agents still adjudicate conformance to architecture and coding
policy.

\textbf{Testing and deployment.}
Modules are integrated, exercised through planned test levels (\emph{e.g.}
system testing, acceptance testing alongside alpha/beta regimes), then released
via deployment (environments, cutover checks, operational readiness).
Agents proliferate synthetic cases, regressions, and deployment manifests; release
gates (sign-off criteria) stay explicit to avoid regressions at scale~\cite{hassan2026trustworthy}.

\textbf{Maintenance.}
Post-release work spans corrective fixes, perfective enhancements, and adaptive
updates for new environments. Agents help triage issues, summarise logs,
propose patches, draft change requests, and keep documentation aligned with drift;
human ownership of production stability is unchanged.

\subsection{Interfaces, Gates, and an End-to-End AI-Assisted Pass}

\textbf{Sequential interfaces.}
Because hand-offs are one-way in the textbook model, mismatches (\emph{e.g.}
SRS statements that designs cannot honour) are cheapest to resolve \emph{before}
downstream artefacts exist. Explicit \textbf{input/output mappings} (\emph{e.g.}
each SDD subsection traces to SRS identifiers; each deployment checklist traces
to tests) tighten these interfaces and complement large-context agentic tooling.

\textbf{Success criteria.}
Per-phase gates (\emph{e.g.\ ``SRS signed''; ``all P1 risks mitigated''; ``CI green
including policy checks''}; \emph{``acceptance tests pass''}; \emph{``runbook \&
rollback verified''}) are the quality backstop replacing informal ``we will fix it
later'' iteration. Agents can propose gate checklists from templates, run static
reviews against specs, or highlight missing traceability;\,\textbf{accountable
humans or independent verification agents} still approve advancement.

\textbf{Illustrative flow.}
A concise agentic analogue of the cascade is: (\emph{i})~\textbf{intent} (fix,
feature, or product);\,
(\emph{ii})~\textbf{requirements synthesis} (\emph{e.g.}
draft SRS and acceptance formulas);\,
(\emph{iii})~\textbf{design and test planning} (\emph{e.g.}
HLD/LLD excerpts, test catalogue);\,
(\emph{iv})~\textbf{development} aligned to SDD and policies;\,
(\emph{v})~\textbf{automated and assisted testing},\ then \textbf{staged deployment};\,
(\emph{vi})~\textbf{observability-informed maintenance} (\emph{e.g.}
clustering incidents, ticketing candidates, speculative improvements from usage
patterns). Sections~\ref{sec:bduf}--\ref{sec:framework} articulate how formal
machine-readable specs and verification loops operationalise such flows at scale.

\subsection{Efficiency Without Abandoning Discipline}

As implementation grows cheaper relative to clarification, organisations can rerun
\textbf{upstream} refinement earlier (smaller batches, shorter frozen windows)
rather than smuggling change only at the code layer. The net effect is a
\textbf{faster} Waterfall-style discipline: the sequence is preserved for
accountability, while AI reduces human effort inside each terrace and strengthens
gates through generated evidence and critiques---mirroring BDUF\,2.0 and SDAD themes
developed next.

\section{The Technical Renaissance: SOTA LLMs \& Long-Context Engineering}
\label{sec:technical}

\subsection{State-of-the-Art Scale Without Published Parameter Counts}

Scale unlocks qualitative capability. Leading vendors rarely disclose exact
parameter counts for flagship chat and coding models, yet frontier systems are
widely understood to occupy the high tens-to-hundreds of billions of trainable
parameters, with the strongest offerings plausibly exceeding \textbf{500B}
parameters once mixture-of-experts routing, tool adapters, and multimodal heads
are counted. We use the term \emph{SOTA LLM} for this class of model; it avoids
tying our claims to any single undisclosed figure (as with many closed
productized systems). In practice, {SOTA LLMs} exhibit reliable architectural
pattern recognition, multi-step logical deduction, and zero-shot generalisation
to novel engineering domains: they recognise design patterns across diverse
frameworks, infer implicit requirements from partial specifications, and
generate architecturally consistent code spanning multiple abstraction layers.

\subsection{Long Context in Practice---and Laboratory ``Megatoken'' Regimes}

Frontier long-context research is often illustrated with extreme benchmarks: the
Gemini~1.5 report, for instance, documents near-perfect long-sequence retrieval
at up to 10 million tokens in controlled evaluations~\cite{geminiteam2024gemini15}.
Day-to-day agentic engineering, however, frequently relies on more modest but
still large windows---on the order of 200k--300k tokens---together with careful
packaging of specifications, dependency summaries, and file slices. Practitioners
routinely obtain strong repository-local results at these scales without placing
every source line of a monorepo into one prompt.

Token-to-line heuristics are illustrative rather than exact (density varies by
language, comments, and formatting), but they clarify orders of magnitude. A
mid-hundreds-of-thousands-token budget already spans a substantial multi-package
service or several tightly coupled repositories. Repositories of about
\textbf{500{,}000 lines of code ({KLOC})} are, by current standards, a modest
size for end-to-end agent-assisted navigation and multi-file refactoring when
process and specs are sound. Monorepos in the \textbf{1--2M {KLOC}} band remain
tractable for {SOTA LLM} workflows, but primarily as an \emph{orchestration}
challenge---combining embeddings, symbol indexes, planning passes, and
tool-using subagents---rather than as a single flat ``ingest the entire tree''
invocation~\cite{otoum2026methods}. That distinction matters: raw context growth
helps, yet the binding constraint increasingly shifts to specification quality,
retrieval policy, and verification---themes developed below.

\subsection{A Real-World Anchor: Claude Code (March 2026 Disclosure)}

In March~2026, reporting on an accidental publication of a JavaScript source map
inside the public \texttt{npm} package for Anthropic's Claude~Code CLI
documents roughly \textbf{512{,}000} lines of TypeScript across almost two
thousand files~\cite{xda2026claudecodeleak}. The same analyses note that the
disclosed architecture centres on persistent and background agents (notably
\textit{Kairos} and memory consolidation via \textit{Auto Dream}), which
third-party commentary reads as heavy internal reliance on the same agentic
programming model the product ships to customers~\cite{xda2026claudecodeleak}.
We do not treat press coverage or user forums as a calibrated measure of
``percentage {AI} authorship''; nothing in the public record replaces an audited
labour study. The episode is nonetheless a useful concrete datum: a
production-grade agentic coding product can reach mid-hundreds-of-{KLOC} scale
while structurally embodying agent-first subsystems.

\subsection{Zero-Shot Repository Synthesis}

Full-project ingestion enables what we term \emph{Zero-Shot Repository
Synthesis}: the ability to generate complex, semantically consistent,
cross-module features from a specification, without losing state across file
boundaries. This effectively eliminates the need for the iterative ``sprinting''
paradigm within the code execution phase of development, as an agent can
synthesise an entire feature branch---including unit tests, integration
scaffolding, and documentation---in a single agentic invocation. The ``AgenticTyper''
case study~\cite{pohle2026agentictyper} demonstrates this capability on
repositories of 81,000+ lines of code.

\subsection{Code Refactoring with {LLMs}}
\label{subsec:llm-refactoring}

Beyond greenfield synthesis, frontier models increasingly support
\emph{refactoring}: restructuring code to improve internal quality---cohesion,
coupling, complexity, smell reduction---while preserving observable behaviour.
That task is technically adjacent to multi-file edits under repository context,
yet it amplifies correctness risk because silent semantic drift is costly.

Recent empirical work contrasts models on open-source Java systems under static
metrics, unit tests, and smell-oriented outcomes~\cite{cordeiro2026tosem}.
Strong proprietary-class models achieve high multifile Pass-at-5-style test
success, while comparative strengths diverge (\emph{e.g.}, smell reduction versus
coupling/cohesion gains). Prompting strategy matters materially: chain-of-thought
and one-shot formulations can widen the refactoring repertoire and improve test
survival versus bare zero-shot prompting. At the same time, developers retain
advantages on refactorings that demand deep, context-sensitive encapsulation or
architecture-level judgement~\cite{cordeiro2026tosem}. A complementary
position paper frames {LLM}-driven refactoring as effective in principle for
assistive tooling, but limited for fully unattended {IDE}-pipeline integration
until hallucination and error-proneness can be bounded; human verification and
research on reliable pipelines therefore remain central~\cite{cordeiro2025ide}.

Within an {SDAD}-style posture, those findings motivate treating {LLM}
refactorings as \textbf{guided transformations}: intent and invariants belong in
the specification and test suites, deterministic checks gate merges, and
specialists adjudicate refactorings whose rationale exceeds what can be anchored
to explicit contracts---mirroring themes of ambiguity control and reviewer
economics argued elsewhere (Figure~\ref{fig:ambiguity}, Section~\ref{sec:bduf}).

\subsection{Code Migration with {LLMs}}
\label{subsec:llm-migration}

\emph{Code migration} denotes repository-scale evolution that retargets languages
or runtimes, moves between frameworks, or upgrades dependencies with
wide blast radius---work that is tedious, often unrewarding, and historically
slow enough to span years when done by hand~\cite{ziftci2025}. It is adjacent to
refactoring (Section~\ref{subsec:llm-refactoring}) but typically imposes stronger
compatibility and rollout constraints because external ecosystems, build graphs,
and organisational queues of pending edits move together.

\subsubsection*{Automation at industrial scale}

A twelve-month, multi-migration case study at Google combined \emph{change
location discovery} with an {LLM} so that developers could steer automation while
retaining responsibility for integration~\cite{ziftci2025}. Across
39~migrations, three developers landed 595~code changes comprising
93{,}574~recorded edits, with the {LLM} authoring roughly three quarters of the
changes and just under seven tenths of the edits; participants reported high
satisfaction and roughly a \textbf{50\%} reduction in elapsed migration effort
relative to prior manual practice~\cite{ziftci2025}. The pattern is instructive for
{SDAD}: the model accelerates mechanical diff generation, while humans supply
programme intent, prioritisation, and integration judgement---the same division of
labour implied by spec-driven gates, but applied to long-lived maintenance
programmes rather than greenfield feature batches.

\subsubsection*{Human--{AI} partnership and trust}

Semi-structured interviews with engineers using an {LLM}-backed migration tool
(Amazon~Q Code~Transformation) surface roles that go beyond passive approval:
participants acted as \emph{directors} (scoping, prompting, sequencing work) and
\emph{reviewers} (vetting model output against project constraints), and they
reasoned about \emph{when} to trust suggestions depending on observable model
outcomes~\cite{omidvartehrani2024}. That evidence reinforces a broader lesson
for agentic pipelines: migration automation is not only a model-quality problem
but a partnership-design problem---tooling must make oversight lightweight enough
that throughput gains survive contact with real governance.

\paragraph{Implications for spec-driven agentic workflows.}

In {SDAD} terms, migrations benefit from treating compatibility rules,
dependency matrices, and verification commands as first-class specification
artefacts: they give both humans and {LLMs} a shared contract for ``done,'' much
as refactoring benefits from tests and invariants (Section~\ref{subsec:llm-refactoring}).
The industrial and {HCI} studies cited here suggest that investments in those
artefacts pay twice---they bound model behaviour and they clarify the trust
transitions developers need when adopting stochastic assistants at
repository scale~\cite{ziftci2025,omidvartehrani2024}.

\subsection{The Ambiguity Tax: Exponential Cost of Vague Specifications}

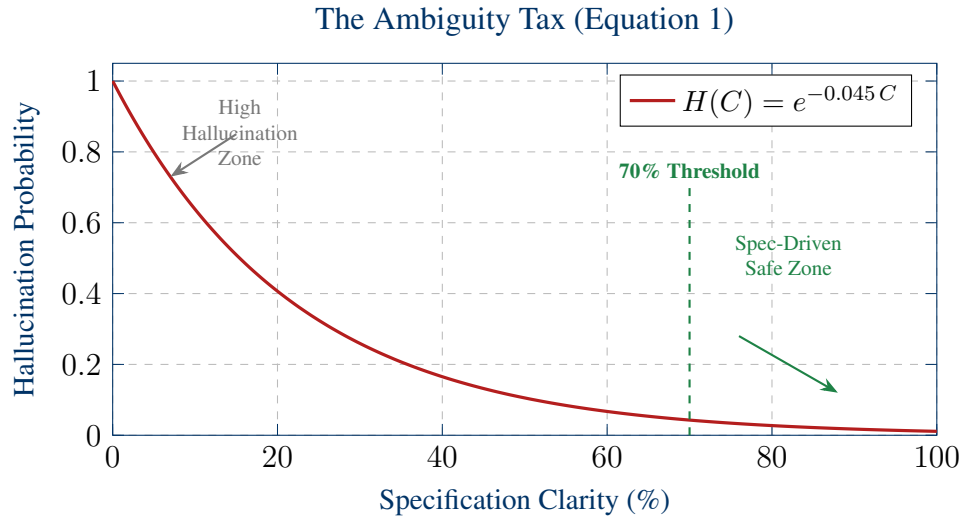
\begin{figure}[htbp]
  \centering
  \begin{tikzpicture}
    \begin{axis}[
      width=0.78\textwidth,
      height=6.5cm,
      xlabel={Specification Clarity (\%)},
      ylabel={Hallucination Probability},
      xmin=0, xmax=100,
      ymin=0, ymax=1.05,
      xtick={0,20,40,60,80,100},
      ytick={0,0.2,0.4,0.6,0.8,1.0},
      yticklabel style={/pgf/number format/fixed, /pgf/number format/precision=1},
      grid=major,
      grid style={dashed, medgray!50},
      axis line style={darkblue},
      tick style={darkblue},
      label style={darkblue, font=\small},
      title style={darkblue},
      title={The Ambiguity Tax (Equation~\ref{eq:ambiguity_tax})},
      legend pos=north east,
      legend style={font=\small}
    ]
      \addplot[very thick, accentred, domain=0:100, samples=200]
        {exp(-0.045*x)};
      \addlegendentry{$H(C) = e^{-0.045\,C}$}

      \draw[-{Stealth}, thick, medgray]
        (axis cs:15,0.85) -- (axis cs:7,0.73)
        node[above right, font=\scriptsize, align=center]
          {High\\Hallucination\\Zone};

      \draw[-{Stealth}, thick, accentgreen]
        (axis cs:76,0.28) -- (axis cs:88,0.12);
      \node[
        accentgreen,
        font=\scriptsize,
        anchor=south,
        align=center,
      ] at (axis cs:82,0.42)
        {Spec-Driven\\Safe Zone};

      \addplot[dashed, thick, accentgreen, domain=0:0.72]
        coordinates {(70,0) (70,0.72)};
      \node[
        font=\scriptsize,
        accentgreen,
        anchor=south,
      ] at (axis cs:70,0.695)
        {\textbf{70\% Threshold}};
    \end{axis}
  \end{tikzpicture}
  \caption{The Ambiguity Tax: as specification clarity decreases toward zero,
    the expected agent hallucination probability rises exponentially.
    Empirical data from~\cite{he2026speed} is consistent with an exponential
    decay model (Equation~\ref{eq:ambiguity_tax}).}
  \label{fig:ambiguity}
\end{figure}

Figure~\ref{fig:ambiguity} illustrates a critical empirical regularity: as
specification clarity decreases, the probability of agent hallucination
increases super-linearly. We term this relationship the \emph{Ambiguity Tax}.
Formally, if $C \in [0,1]$ denotes specification clarity and
$H(C)$ denotes the expected hallucination probability, empirical data is
consistent with:

\begin{equation}
  H(C) \approx \alpha \cdot e^{-\beta C}, \quad \alpha,\beta > 0
  \label{eq:ambiguity_tax}
\end{equation}

where $\alpha$ represents baseline hallucination probability at zero clarity
and $\beta$ governs the rate at which clarity suppresses hallucination.

\section{AI-Code: A Fourth Paradigm of Software Production}
\label{sec:ai-code}

Software has historically been produced through one of three paradigms that differ
primarily on \emph{who writes the code} and at what \emph{skill barrier}.
\emph{Pro-code} denotes hand-authored source written by professional developers in
standard programming languages, where every line is traceable to a named human
commit. \emph{Low-code} covers visual, model-driven platforms---such as Microsoft
Power Apps or OutSystems---that require limited scripting and offer medium governance
coverage, with customisation zones that can escape IT controls. \emph{No-code}
describes purely graphical tools (e.g.\ Webflow, Zapier) accessible to business
users with no programming knowledge, but at the cost of low governance visibility
and significant shadow-IT proliferation risk~\cite{jain2026agentic}.

This paper introduces a fourth paradigm: \emph{AI-code}, defined as
\textbf{source code or configuration artefacts generated by an AI model or agentic
system, regardless of which interface---pro-code {IDE}, low-code platform, or
no-code tool---the human used to invoke it}. A developer using {Claude Code} or
{GitHub Copilot} agent is producing AI-code via a pro-code shell; a product manager
using Replit or Lovable is producing AI-code via a no-code shell; a business analyst
invoking Copilot Studio is producing AI-code via a low-code shell. In all three
cases the underlying governance risk profile---ungoverned AI-generated logic, opaque
provenance, and potential architecture sprawl---is structurally
identical~\cite{jain2026agentic}.

\paragraph{What makes AI-code categorically different.}
The first three paradigms differ on who writes the code and at what skill barrier.
AI-code changes the question entirely. Three structural properties distinguish it
from all prior paradigms.

\begin{enumerate}[noitemsep]
  \item \textbf{Author--accountable decoupling.} The entity that generates the code
    (the model) cannot be held legally or professionally accountable. Accountability
    reverts to the human who issued the prompt or approved the output---roles that
    most organisations have not yet formally defined~\cite{hassan2026trustworthy}.
  \item \textbf{Skill-independent expressiveness.} A no-code user can now generate
    production-grade backend logic by prompting an agent. The quality ceiling of
    no-code effectively disappears; the governance risk ceiling rises to match
    pro-code~\cite{jain2026agentic}.
  \item \textbf{Invisible paradigm convergence.} Pro-, low-, and no-code produce
    distinct artefact types that IT can audit separately. AI-code can produce
    \emph{any} artefact type from any entry point, making categorical auditing
    impossible without provenance instrumentation at the generation
    layer~\cite{he2026speed,jain2026agentic}.
\end{enumerate}

\paragraph{The accountability question.}
For traditional code---pro-, low-, or no-code---the governance question is
\emph{who wrote it?} For AI-code, the question shifts: \emph{who specified it, who
reviewed it, and who approved the merge?} This shift has a direct methodological
implication. Because ambiguity in the specification is amplified rather than
compensated by an AI agent~\cite{he2026speed}, the quality of human intent becomes
the primary determinant of AI-code quality. These are precisely the responsibilities
that {SDAD}'s Spec Architect role, formal specification gates, and provenance
logging are designed to make explicit and auditable
(Section~\ref{sec:framework}).

\section{The ``Waterfall Resurgence'': Big Design Up Front (BDUF) 2.0}
\label{sec:bduf}

\subsection{Context Ingestion as Execution}

Modern agents utilise massive context windows to process Functional Requirement
Documents (FRDs) and use-case lists simultaneously. Because an agent can
generate a full-stack project from a specification in hours rather than months,
the traditional Waterfall risk of obsolescence---specifications becoming stale
before implementation is complete---is substantially mitigated. High-fidelity
specifications have become the primary \emph{execution fuel} for {SOTA LLMs}:
the richer and more precise the specification, the faster and more correct the
resulting synthesised code~\cite{otoum2026methods}.

\subsection{Implication for BDUF: Clarity as an Execution Constraint}

As formalised by Equation~\ref{eq:ambiguity_tax} and illustrated in
Figure~\ref{fig:ambiguity}, hallucination risk increases non-linearly as
specification clarity declines. In BDUF-style agentic delivery, large context
windows do not remove that penalty; they often increase the blast radius when a
single vague requirement propagates inconsistent edits across many files.
Consequently, teams benefit from concise, testable requirement artefacts before
long synthesis passes~\cite{he2026speed}.

\subsection{Agentic Linearism}

We introduce the term \emph{Agentic Linearism} to describe the empirically
observed efficiency advantage of linear, ``one-shot'' builds for enterprise-
level modules. Unlike iterative human development, which benefits from
incremental feedback and course-correction, an agentic pipeline operating from
a complete and unambiguous specification is most efficient when executed as a
single, non-interrupted synthesis pass. Interruptions, partial specifications,
or mid-synthesis requirement changes introduce context fragmentation that
degrades output quality. This observation rehabilitates the Waterfall principle
of ``requirements freeze'' for the agentic execution phase---not as bureaucratic
inertia, but as a technical optimisation.

\section{The Metamorphosis of Agile: From Sprints to Continuous Synthesis}
\label{sec:agile}

The transition from purely human Scrum to agent-augmented practice is clearer if
the canonical Sprint rhythm is pinned first. Scrum organises iterative delivery
into a Sprint time-box with four events alongside three artefacts and their
paired commitments (\emph{e.g.\ Product Goal},\,\emph{Sprint Goal},\,\emph{Definition
of Done}); the empirical inspect--adapt cadence survives even when tooling changes.
Coding assistants chiefly
reallocate \emph{where} throughput appears (faster backlog refinement,
implementation parallelism, churn in generated artefacts) versus \emph{where}
attention must harden into human review gates (acceptance judgments, architecture
risk, ambiguity in frontier-model behaviour). Frontier agents operate as tooling
\textbf{within} Developers' accountability; they should not appear as informal
replacement for Product Owner priorities or Scrum Master facilitation.
Table~\ref{tab:scrum-sdad-shift} restates the same Sprint-centred vocabulary
alongside SDAD-aligned names and emphasis: human-led inspect--adapt ceremonies
remain where judgment, negotiation, and reflection live, while agents accelerate
backlog narration, implementation, and test execution under explicit review gates.
Product ownership and feasibility decisions stay human; orchestration of model
capabilities and prompt--review debt may concentrate in roles variously labelled
\emph{AI coordinator} or \emph{AI supervisor} when implementers are predominantly
agentic, even as a human facilitator still upholds transparency and flow for the
mixed team.

\begin{table}[htbp]
  \centering
  \caption{Scrum artefacts and events mapped to SDAD-oriented practice. Wording stresses continuous synthesis and value delivery; where noted, ceremonies remain human discussions that may be informed by machine-generated analytics.}
  \label{tab:scrum-sdad-shift}
  \footnotesize
  \renewcommand{\arraystretch}{1.22}
  \setlength{\tabcolsep}{3.5pt}%
  \begin{tabular}{@{}%
    >{\bfseries\raggedright\arraybackslash}p{0.23\linewidth}
    >{\raggedright\arraybackslash}p{0.74\linewidth}@{}}
    \toprule
    Scrum (human-agile) &
    SDAD-oriented counterpart and notes \\
    \midrule
    Sprint &
      \textbf{Development cycle} or \textbf{iteration}: a time-box for delivery,
      reframed toward continuous synthesis rather than a brittle boundary that
      interrupts flow. \\
    Sprint Planning &
      \textbf{Iteration planning} / \textbf{development-cycle planning}: select
      goals and tasks for the upcoming cycle, including acceptance scaffolding
      and scope judgment assisted---not replaced---by backlog narration tools. \\
    Daily Scrum &
      Not intrinsic to automated pipelines; short \textbf{human stand-ups} remain
      useful for blockers, dependencies, and escalation when people and agents
      share accountability. \\
    Sprint Review &
      \textbf{Development-cycle review} / \textbf{iteration review}: inspect the
      increment, gather stakeholder feedback, and reserve explicit human capacity
      for acceptance and risk calls. \\
    Sprint Retrospective &
      \textbf{Development-cycle retrospective} / \textbf{iteration retrospective}:
      team reflection and improvement actions; AI may surface metrics or
      suggestions, but decisions and norms are owned by people. \\
    Increment &
      \textbf{Deliverable} / \textbf{release}: the usable outcome of a cycle,
      stressing customer value; AI-assisted build and test can shorten the path to
      a shippable increment when specification fidelity and review keep pace. \\
    Sprint Backlog &
      \textbf{Iteration backlog}: committed work for the cycle, often including
      defect triage, technical debt, and refactor tasks surfaced during
      agent-assisted development. \\
    Product Backlog &
      \textbf{Project} / \textbf{feature backlog}: ordered intake for successive
      cycles, continuously refined from feedback, telemetry, and changing
      requirements---with machine help on drafting or decomposing items under
      Product Owner direction. \\
    \bottomrule
  \end{tabular}
\end{table}

\begin{figure}[htbp]
  \centering
  \includegraphics[width=\linewidth]{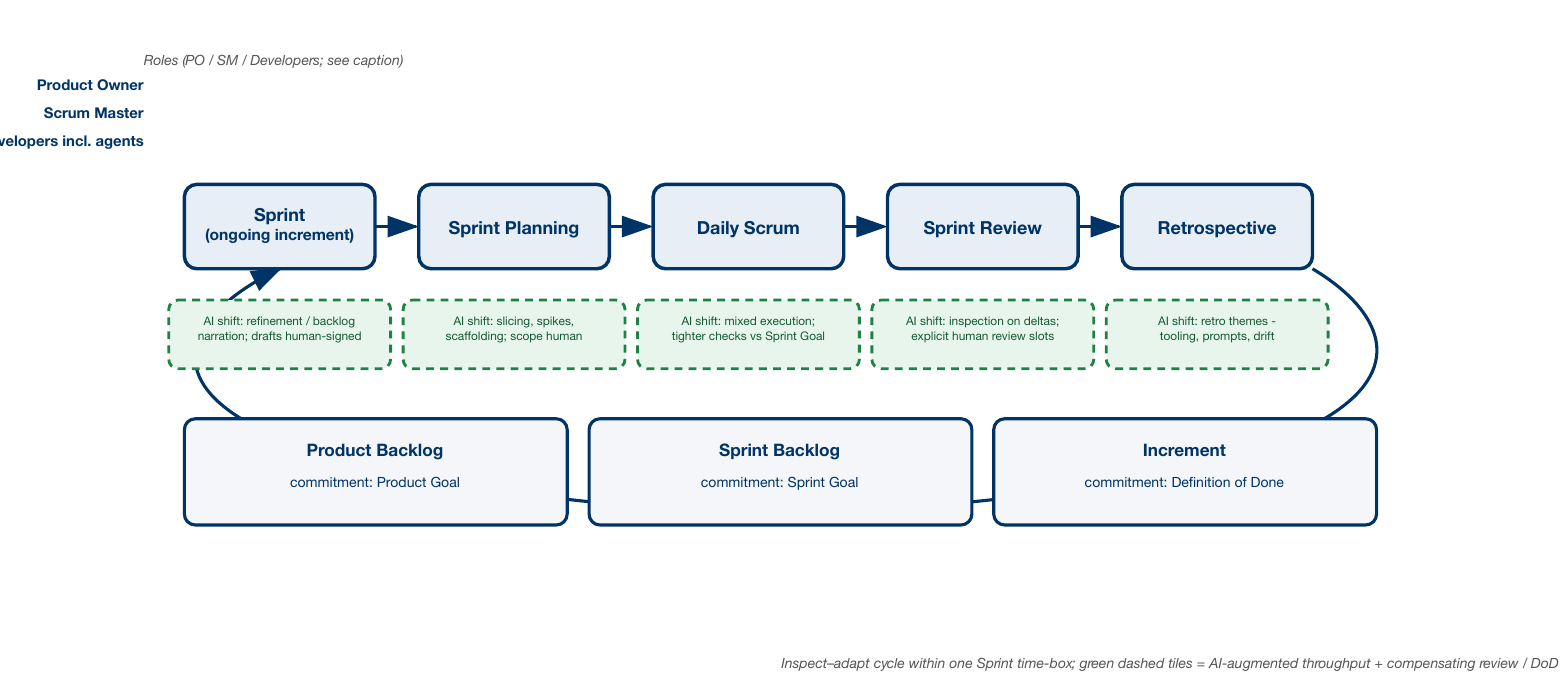}
  \caption{Scrum within one \emph{Sprint}: an inspect--adapt loop across \emph{Sprint Planning}, \emph{Daily Scrum}, \emph{Sprint Review}, and \emph{Sprint Retrospective} (heavy arrows), with \emph{Product Backlog} (Product Goal), \emph{Sprint Backlog} (Sprint Goal), and \emph{Increment} (Definition of Done) below. Green dashed tiles mark AI support, while accountability stays human: \emph{Product Owner}, \emph{Scrum Master}, and \emph{Developers}.}
  \label{fig:scrum-ai}
\end{figure}

Figure~\ref{fig:scrum-ai} summarises the same scaffolding as a sprint-centred process diagram; the caption records how \emph{Product Owner}, \emph{Scrum Master}, and \emph{Developers} engage across events when coding assistants sit inside Developers' accountability.

\subsection{The Obsolescence of the 2-Week Sprint}

Even when the scaffolding of Figure~\ref{fig:scrum-ai} persists on paper, the two-week Scrum sprint was calibrated to human cognitive and social rhythms:\ the time required for a small team to design, implement, review, and integrate
a meaningful unit of functionality. In 2026, {SOTA LLMs} compress 80 hours of
development work into minutes~\cite{otoum2026methods}. The sprint boundary---once
a meaningful unit of value delivery---becomes an artificial constraint that
interrupts the continuous synthesis cycle of agentic pipelines.

\subsection{Shift from ``Story Points'' to ``Spec Fidelity''}

In Human-Agile, value was measured in story points---a proxy for human effort.
In Agentic-SDAD, effort is no longer the binding constraint. The binding
constraint is \emph{specification fidelity}: the degree to which a requirement
document unambiguously captures the intended system behaviour. We propose
``Spec Fidelity'' as the primary quality metric of the Agentic SDLC, measured
along four dimensions: \emph{completeness}, meaning that edge cases and failure
modes are explicitly documented; \emph{consistency}, meaning that requirements
do not contain internal contradictions; \emph{unambiguity}, meaning that each
requirement admits a single, stable interpretation; and \emph{verifiability},
meaning that every requirement can be mapped to automated tests and objective
pass/fail criteria.

\subsection{The SDAD-V Model}

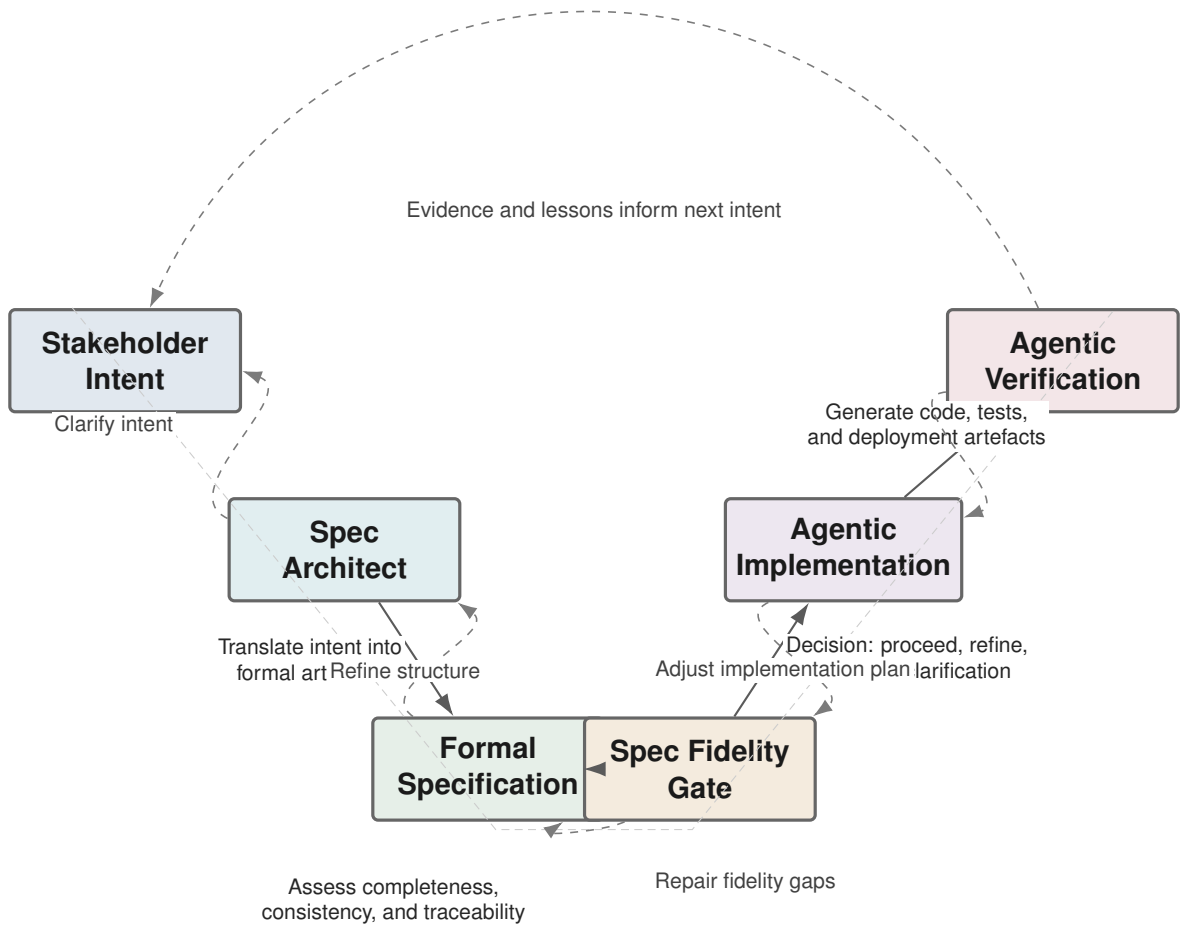
\begin{figure}[htbp]
  \centering
\begin{tikzpicture}[
  stage/.style={
    rectangle,
    rounded corners=2pt,
    minimum width=3.05cm,
    minimum height=1.35cm,
    align=center,
    draw=black!60,
    very thick,
    font=\sffamily\small\bfseries,
    text=black!90,
    fill=white
  },
  fwd/.style={-{Latex[length=3mm,width=2mm]}, thick, draw=flowGray},
  fb/.style={-{Latex[length=2.6mm,width=1.8mm]}, semithick, draw=flowGray!80, dashed},
  fwdlabel/.style={font=\sffamily\scriptsize, text=black!85, align=center, fill=white, inner sep=1pt},
  fblabel/.style={font=\sffamily\scriptsize, text=black!75, align=center, fill=white, inner sep=1pt}
]

\node[stage, fill=stageBlue!14]  (intent)    at (-6.2,  4.2) {Stakeholder\\Intent};
\node[stage, fill=stageTeal!14]  (architect) at (-3.3,  1.7) {Spec\\Architect};
\node[stage, fill=stageGreen!14] (formal)    at (-1.4, -1.2) {Formal\\Specification};
\node[stage, fill=stageAmber!16] (gate)      at ( 1.4, -1.2) {Spec Fidelity\\Gate};
\node[stage, fill=stagePlum!14]  (impl)      at ( 3.3,  1.7) {Agentic\\Implementation};
\node[stage, fill=stageRed!14]   (verify)    at ( 6.2,  4.2) {Agentic\\Verification};

\draw[densely dashed, draw=black!20] (-6.9,4.9) -- (-1.3,-2.0) -- (1.3,-2.0) -- (6.9,4.9);


\draw[fwd] (architect) -- (formal)
  node[midway, left=4pt, fwdlabel] {Translate intent into\\formal artefacts};

\draw[fwd] (formal) -- (gate);
\node[fwdlabel] at ($(formal.south)+(-1.25,-1.05)$) {Assess completeness,\\consistency, and traceability};

\draw[fwd] (gate) -- (impl)
  node[midway, right=4pt, fwdlabel] {Decision: proceed, refine,\\or request clarification};

\draw[fwd] (impl) -- (verify)
  node[midway, above=1pt, xshift=-0.35cm, fwdlabel] {Generate code, tests,\\and deployment artefacts};

\draw[fb] (architect.165) to[out=155, in=-20,  looseness=1.1] (intent.-5);
\draw[fb] (formal.145)    to[out=145, in=-35,  looseness=1.1] (architect.-25);
\draw[fb] (gate.215)      to[out=205, in=-145, looseness=1.0] (formal.-35);
\draw[fb] (impl.215)      to[out=205, in=35,   looseness=1.1] (gate.25);
\draw[fb] (verify.195)    to[out=185, in=20,   looseness=1.1] (impl.15);

\node[fblabel] at (-6.35, 3.35) {Clarify intent};
\node[fblabel] at (-2.5,  0.1)  {Refine structure};
\node[fblabel] at ( 2.0, -2.7)  {Repair fidelity gaps};
\node[fblabel] at ( 2.5,  0.1)  {Adjust implementation plan};

\draw[fb] (verify.115) to[out=115, in=65, looseness=1.25] (intent.65);
\node[fblabel] at (0,6.2) {Evidence and lessons inform next intent};

\end{tikzpicture}
  \caption{The SDAD-V Model. Intent is formalised down the left arm
    (Spec Architect $\to$ Formal Specification $\to$ Fidelity Gate) and
    executed up the right (Agentic Implementation $\to$ Agentic Verification).
    Dashed arrows denote local and system-level feedback loops.}
  \label{fig:cycle}
\end{figure}

Figure~\ref{fig:cycle} depicts the SDAD-V Model, which is structurally
inspired by the classical waterfall V-model. The left branch descends from
stakeholder intent through the Spec Architect to the Formal Specification and
Spec Fidelity Gate at the vertex, representing progressive formalisation of
requirements. The right branch ascends through Agentic Implementation to
Agentic Verification, representing construction and validation against the
formal specification. Solid arrows indicate the primary forward flow of
formalised artefacts, whereas dashed arrows indicate iterative feedback loops
that return verification evidence to earlier stages, including intent-level
refinement.

\section{Proposed Framework: Spec-Driven Agentic Development (SDAD)}
\label{sec:framework}

\subsection{The Role of the Spec Architect: Redefining the Software Engineer}

Within the SDAD framework, the primary human role is that of the
\textbf{Spec Architect}---a practitioner whose value lies not in the ability to
implement code, but in the ability to capture, formalise, and validate
\emph{intent}. The Spec Architect possesses:

\begin{enumerate}[noitemsep]
  \item \textbf{Domain Fluency:} Deep understanding of business logic,
    stakeholder needs, and system constraints.
  \item \textbf{Formal Methods Literacy:} Ability to express requirements in
    machine-interpretable notations (YAML-based logic, structured
    natural language templates).
  \item \textbf{Agentic Orchestration Skill:} Proficiency in designing
    multi-agent pipelines, including verification agents, security-review
    agents, and regression-test agents.
  \item \textbf{Adversarial Reasoning:} Ability to anticipate agentic failure
    modes (hallucination, specification gap exploitation) and design
    specifications that are robust against them.
\end{enumerate}

\subsection{Team Role Metamorphosis: Beyond the Engineer}
\label{subsec:team-metamorphosis}

The Spec Architect role captures the most visible shift in the delivery team,
but agentic AI redistributes responsibility across \emph{every} function.
Roles that were previously peripheral to code production become structurally
central; the change is not one of headcount but of accountability and
decision-making authority~\cite{jain2026agentic}.

\paragraph{Quality Assurance: from test authoring to evaluation strategy.}
In human-centric pipelines, QA practitioners primarily author tests, execute
regression suites, and triage defects. Under SDAD, synthesis agents produce
much of that artefact volume, so QA shifts to defining evaluation policy:
behavioural oracles, coverage contracts, and rollback criteria that determine
what the pipeline is permitted to release.

\paragraph{SRE and platform engineering: from operations execution to
autonomy-control infrastructure.}
Site reliability and platform teams have traditionally enforced discipline
through runbooks, incident response, and deployment pipelines. In SDAD, where
agents can modify release-bound artefacts, the priority becomes the
autonomy-control layer: permission boundaries, observability for agent
behaviour, and preserved human rollback authority, with traceability and audit
trails becoming decisive at production proximity~\cite{jain2026agentic}.

\paragraph{Product owner and business analyst: from backlog wording to
intent precision.}
Agile delivery often tolerates informal disambiguation, with ambiguous backlog
items resolved through conversation across product, engineering, and QA\@.
Agentic pipelines execute intent as written, so product owners and business
analysts must front-load precision by expressing business intent as
unambiguous, edge-case-aware acceptance criteria before synthesis, or risk
strategically incorrect artefacts that are only caught at verification
gates~\cite{he2026speed}.

Table~\ref{tab:role-metamorphosis} summarises the shift across the full
delivery team alongside the Spec Architect role introduced above.

\begin{table}[htbp]
  \centering
  \caption{Role metamorphosis across the delivery team under SDAD\@. Each
    role shifts from execution-oriented work towards higher-order judgment,
    strategy, and control~\cite{jain2026agentic}.}
  \label{tab:role-metamorphosis}
  \renewcommand{\arraystretch}{1.28}
  \begin{tabularx}{\textwidth}{%
    >{\bfseries\raggedright\arraybackslash}p{2.9cm}
    >{\raggedright\arraybackslash}X
    >{\raggedright\arraybackslash}X}
    \toprule
    \textbf{Role} & \textbf{Human-Agile focus} & \textbf{SDAD focus} \\
    \midrule
    Spec Architect (Engineer) &
      Code authorship, PR review, refactoring &
      Intent capture, formal specification, agentic orchestration \\
    QA / Test Engineer &
      Test-case authoring, regression execution &
      Evaluation strategy, behavioural oracle design, coverage contracts \\
    SRE / Platform &
      Runbook execution, deployment pipelines, incident response &
      Autonomy-boundary definition, observability infrastructure,
      rollback authority \\
    Product Owner / BA &
      Backlog wording, sprint planning, stakeholder liaison &
      Intent precision, edge-case elicitation,
      acceptance-criteria accountability \\
    \bottomrule
  \end{tabularx}
\end{table}

\subsection{The SDAD Workflow}

The SDAD workflow comprises four principal phases. First, \emph{intent capture}
uses structured human--agent dialogue to elicit requirements, constraints, and
business rules into a reviewable intent record. Second, \emph{formal
specification} converts that record into a machine-interpretable blueprint that
defines interfaces, logic, data structures, and acceptance criteria~\cite{otoum2026methods}.
Third, \emph{agentic synthesis} treats the specification as the authoritative
input for multi-file generation of implementation artefacts and supporting
tests~\cite{pohle2026agentictyper}. Fourth, \emph{independent verification}
checks generated outputs against the specification through automated test,
security, and conformance gates before release. The distinguishing governance
constraint is that merge or release authority remains human: a Spec Architect or
equivalent accountable reviewer signs off after verification rather than
delegating final responsibility to the synthesis agents.

\subsection{Automated testing in the AI-native toolchain}
\label{subsec:ai-testing}

\textbf{Placement and intent.}
Automated assurance sits at two coupled interfaces in agentic pipelines: artefacts
\textbf{exercising the product under test}, and artefacts \textbf{critiquing outputs
of synthesis agents}. This subsection summarises how industrial test automation,
{CI}/{CD} practice, and the research literature align with SDAD verification
gates (Steps~3--4 above).

\subsubsection{Industrial {AI}-{ML}-augmented end-to-end automation}

Classical scripted automation trims latency on stable systems, yet under frequent
iteration the same selectors, flow assumptions, and oracles degrade: maintenance
consumes disproportionate effort and brittle suites stall pipelines~\cite{pham2022}.
Surveys of end-to-end test automation tooling show {AI}/{ML} being applied across
\textbf{generation}, \textbf{adaptation}/{self-healing}, \textbf{bucketing and
prioritisation}, \textbf{flaky-test analytics}, \textbf{coverage-guided
exploration}, and allied activities---explicitly tying techniques to tooling
capabilities so practitioners can relate research claims to shipped
products~\cite{pham2022}. For SDAD, that mapping supports a pragmatic rule:
accelerate authoring and upkeep with assistants, but keep \emph{release authority}
outside the same model family that authored or repaired tests unless redundancy
(for example independent agents, deterministic property checks, or human sign-off)
is designed in~\cite{hassan2026trustworthy}.

\subsubsection{Large language models on core testing tasks}

Pre-trained large language models are experimentally pervasive across downstream
testing work---generation and augmentation of suites, oracle and input synthesis,
repair and debugging scaffolding, often adjacent to metamorphic or search-based
baselines~\cite{wang2024}. \citet{wang2024} systematically review more than one
hundred studies and consolidate task taxonomies (\emph{e.g.,} test-case preparation
versus program repair), model and prompt regimes, datasets, metrics, and open
research gaps; the field reads as heterogeneous and fast-moving rather than
settled behind a dominant recipe.

\subsubsection{Narrow-scope empirical contrasts}

Controlled studies emphasise how much \emph{context engineering} dominates
reported usefulness. Comparative web-form generation across eleven models and three
HTML-oriented prompt-shaping strategies exposes wide spreads in whether generated
interaction scripts actually submit cleanly; tighter structural prompts that
expose form semantics consistently outperform raw page dumps~\cite{li2026}.
Separately, a multi-project {Java} study spanning open-weight code models through
commercial {GPT-4}, with search-based \texttt{EvoSuite} as a classical reference,
reports strong sensitivity of coverage and mutation outcomes to prompting and
few-shot framing, and underscores that frontier {LLMs} remain complementary rather
than drop-in substitutes for disciplined search~\cite{yang2024}.

\subsubsection{Reading across to {SDAD} Step~{4}}

In sum, tooling surveys foreground \textbf{adaptation pressure in {CI}/{CD}}, broad
reviews foreground \textbf{capability dispersion across tasks}, and focused
benchmarks foreground \textbf{oracle quality and contextual prompts}~\cite{pham2022,wang2024,li2026,yang2024}.
These strands jointly reinforce SDAD\@: autonomous test synthesis is valuable
evidence only when embedded in workflows where specifications, deterministic
checks, redundancy among agents or tools, and human gatekeeping jointly remain
authority over merge and release~\cite{hassan2026trustworthy}.

\section{Comparative Analysis: SDLC 2020 vs.\ SDLC 2026}
\label{sec:comparison}

Table~\ref{tab:comparison} summarises the principal differences between
classical Waterfall, the Human-Agile methodology prevalent in 2020, and the
Agentic-SDAD paradigm of 2026.

\begin{table}[htbp]
  \centering
  \setlength{\tabcolsep}{4pt}%
  {\scriptsize\caption{Waterfall (1970s), Human-Agile / Scrum (2020), and Agentic-SDAD (2026)}}
  \label{tab:comparison}
  \footnotesize
  \renewcommand{\arraystretch}{1.28}
  \begin{tabularx}{\textwidth}{%
    >{\bfseries\raggedright\arraybackslash}p{2.55cm}
    >{\raggedright\arraybackslash}X
    >{\raggedright\arraybackslash}X
    >{\raggedright\arraybackslash}X}
    \toprule
    Dimension &
    \textbf{Waterfall (1970s)} &
    \textbf{Human-Agile / Scrum (2020)} &
    \textbf{Agentic-SDAD (2026)} \\
    \midrule
    Primary Artefact &
      Signed-off specs.\ \& baseline designs &
      Working Software &
      Machine-Readable Specification \\
    Core Human Skill &
      Analysis, documentation \& gatekeeping &
      Full-Stack Programming &
      Spec-Driven Architecture \\
    Bottleneck &
      Change control \& sequential hand-offs &
      Developer Headcount &
      Logic Clarity \& Context Quality \\
    Feedback Loop &
      Months (phase reviews) &
      Weeks (Sprint Review) &
      Minutes (Agentic Test Loops) \\
    Documentation &
      Comprehensive, phase-delivered &
      ``Just Enough'' (Agile Manifesto) &
      Comprehensive \& Formal \\
    Sprint Unit &
      Sequential phases / milestones &
      2-Week Sprint &
      Continuous Synthesis Cycle \\
    Velocity Metric &
      Schedule \& scope adherence &
      Story Points (Human Effort) &
      Spec Fidelity Score \\
    Code Authorship &
      Human-Written &
      Human-Written &
      Agent-Synthesised \\
    Technical Debt Risk &
      Late discovery at integration &
      Managed via Refactoring Sprints &
      Mitigated via Specification Gates~\cite{he2026speed} \\
    Security Review &
      Formal gate reviews \& checklists &
      Manual Penetration Testing &
      Automated SAST / Agent Security Gates \\
    \bottomrule
  \end{tabularx}
\end{table}

\subsection{Velocity and Quality Metrics}

This comparison should be read as directional evidence rather than a universal
causal guarantee: published results vary by task mix, codebase maturity, and
governance regime. Empirical data from \citet{he2026speed} documents a
3--5$\times$ velocity gain for teams adopting AI-assisted coding; in the same
study, weak requirement discipline is associated with higher technical debt
(cyclomatic complexity, code duplication, and test coverage degradation), while
rigorous requirement documentation is associated with substantially smaller
quality trade-offs. Accordingly, SDAD is framed here as reducing---not
eliminating---the velocity/quality tension via formal specifications,
verification gates, and HITL sign-off.
Section~\ref{sec:sdad-metrics} situates these shifts within a broader taxonomy
of Waterfall, Scrum, and SDAD measurement lenses.

\subsection{Communication hierarchies, handoffs, and collapse under agentic specification}
\label{subsec:communication-hierarchy}

Human-centric Agile delivery presumes a \emph{relay} of intent through role
specialisation: business stakeholders brief a product owner; the product owner
prioritises backlog items for an architect or tech lead; senior engineers shard
work for junior implementers; quality and operations consume partially
reconstructed rationale at each boundary. Each hop introduces latency,
compression, and tacit reinterpretation---what we might call a cumulative
\textbf{translation tax}: subtle constraints and acceptance semantics drift as
natural-language summaries move down the hierarchy.

This pattern is not accidental. Classic organisational theory already ties system
structure to the communication graph of the designing organisation~\cite{conway1968},
and software-project folklore quantifies why purely horizontal scaling of
implementers hits diminishing returns as pairwise coordination explodes~\cite{brooks1995}.
Layered SDLC roles are partly an \emph{economic response} to bounded human
attention and serialisation of expertise.

\textbf{Agentic SDAD can change the shape of that graph.} Frontier models can hold
long-form requirements and generated artefacts in a shared context window at
once~\cite{geminiteam2024gemini15}, and multi-agent pipelines can internalise
activities that previously sat in separate human queues~\cite{otoum2026methods}.
In limit form, a machine-readable Specification acts as a single source of
truth: coding can become less of a separate phase of serial reinterpretation and
becomes the terminal rendering of intent already expressed at high fidelity---the
number of serial human handoffs can be reduced, with coordination shifting toward
\emph{specification fidelity}, verification gates, and orchestration policy.

LLMs can function here as \textbf{broad translators} between natural-language
intent (including stakeholder prose refined into structured specs) and
machine-level implementation~\cite{salem2024survey}. That shift explains why
marginal implementation effort may fall toward commodity inference spend
(Section~\ref{sec:economics}) while the scarce inputs migrate upstream to logic
clarity and gate design~\cite{he2026speed,hassan2026trustworthy}.

SDAD nevertheless deliberately targets a \textbf{mixed} regime: the Spec
Architect remains accountable for domain correctness, adversarial specification,
and human-in-the-loop sign-off (Section~\ref{sec:framework}). Fully autonomous,
human-absent pipelines may become feasible for narrow domains, but they amplify
risks around hallucination, security, and operational comprehension---trade-offs
surveyed across agentic software-engineering practice~\cite{otoum2026methods,salem2024survey}.
Recent systems research on configurable human--AI software teams illustrates how
agents can adopt differentiated roles while humans participate as peers rather
than only as passive reviewers~\cite{klieger2024chatcollab}, which aligns with
SDAD's orchestrated synthesis plus verification loops rather than unchecked
single-agent codegen.

\section{Metrics for the SDAD Era}
\label{sec:sdad-metrics}

Historical \textbf{Software Engineering~1.0} (Waterfall) privileged
\textbf{volume} proxies---thousands of lines of code (KLOC) and function
points---because labour-intensive specification and hand-coding made output
scale a natural management lens. \textbf{Software Engineering~2.0}
(Agile/Scrum) shifted attention to \textbf{iteration speed}: sprint velocity,
burndown charts, and story points calibrated team capacity and flow rather than
semantic alignment with a unitary formal document.

\textbf{Software Engineering~3.0} (SDAD) treats implementation as
increasingly commoditised: synthesis unfolds at inference speed, so raw counts
of lines of code or story points are weak surrogates for product
quality---analogous to judging a vehicle chiefly by paint volume rather than
performance. The operative question is whether the machine faithfully
instantiates \emph{intent}. Organisational metrics must therefore migrate from
\textbf{labor-tracking} to \textbf{intent-validation}.

As Royce~\cite{royce1970} stressed, analysis and coding are essential, yet large
programmes cannot survive as a naive two-step process: the trajectory is ``doomed
to failure'' without the disciplined elaboration he associated with
plan-driven engineering. In the agentic era, that discipline shows up in what we
measure at the ``grandiose'' pole: the specification---machine-readable,
auditable, and dense enough to steer synthesis without chronic repair loops.

Table~\ref{tab:comparison} aligns eras on artefacts and cadence, but the row labels
are \emph{not} dimensionally commensurate across columns: one cannot equate KLOC
with sprint velocity or with the Synthesis Efficiency Ratio
(\textsc{ser}). Instead, each era carries primary indicators tuned to its
bottleneck; juxtaposing them illustrates how methodological shifts relocate
scarcity---from coding throughput (1.0), through iteration mechanics (2.0), to
specification fidelity and orchestration (3.0). Table~\ref{tab:metric-evolution}
summarises that evolution at a glance; subsections below expand representative
3.0 metrics that complement the economic formalism in
Section~\ref{sec:economics}.

\subsection{Synthesis Efficiency Ratio (\textsc{ser})}

The \textsc{ser} is the lead key performance indicator for the Spec Architect:
it scores how
compactly a high-page-count \textsc{frd} (the encoded intent) converts into
working code without wasteful agentic loop-back or hallucinated scaffolding. A
high \textsc{ser} signals deterministic, high-fidelity specification and
near--one-shot synthesis; a low \textsc{ser} signals a heavy \textbf{ambiguity
tax}---vague requirements force the agent to iterate, consume additional tokens,
and guess latent architectural commitments. Operationally, \textsc{ser} is
consistent with the tokenised cost view developed below; see
Equation~\eqref{eq:ser} and the role of the repair multiplier~$\varphi$ in
Equation~\eqref{eq:tci-agentic}.

\subsection{Spec fidelity and ambiguity tax}

Where Scrum emphasised \textbf{velocity} (how fast the team moves cadence), SDAD
emphasises \textbf{fidelity} (how accurately the machine consumes the
specification). Two illustrative observables are:
\begin{itemize}[leftmargin=*,nosep]
  \item \textbf{Ambiguity tax:} the gap between expected token spend for a
    feature slice (given a crisp, testable \textsc{frd}) and realised spend when
    prose-heavy or under-specified fragments force clarification prompts and
    retries, tied to the structural view in Equation~\eqref{eq:ambiguity_tax}.
  \item \textbf{First-pass alignment rate:} the share of requirements whose
    first synthesis pass already satisfies automated checks and human acceptance
    semantics without material rework.
\end{itemize}

\subsection{Marginal cost of code and inference economics}

When loaded human \emph{man-months} cease to dominate the variable cost of
delivery, leaders must instrument \textbf{inference economics} alongside residual
architecture labour:
\begin{itemize}[leftmargin=*,nosep]
  \item \textbf{Cost per implementation (\textsc{cpi}):} combined token invoice
    at effective $P_{\mathrm{token}}$ plus accountable Spec Architect and
    gatekeeper time (the additive human term in Equation~\eqref{eq:tci-agentic}).
  \item \textbf{Re-synthesis return:} comparative cost of human patch-oriented
    repair versus regenerating a module from an updated specification when the
    fault traces to intent drift rather than a local code defect---the strategic
    complement to the \textsc{tci} framing in Section~\ref{sec:economics}.
\end{itemize}

\subsection{Agentic Autonomy Rate (\textsc{aar})}

\textsc{aar} summarises how much of the merged codebase originates from governed
agent synthesis rather than bespoke human typing---a practical analogue for how
``hands-off'' continuous synthesis can become while still honouring verification
gates. Colloquial practice sometimes links this ratio to conversational
\emph{vibe coding}; SDAD reframes the same telemetry as an auditable autonomy
metric paired with human touchpoints, not a substitute for
specification-first governance. Table~\ref{tab:aar-metrics} lists companion
indicators teams can trend alongside \textsc{aar}.

\begin{table}[htbp]
  \centering
  \small
  \caption{Representative autonomy and verification indicators for SDAD-style
    delivery (illustrative targets; organisations should calibrate to domain risk).}
  \label{tab:aar-metrics}
  \renewcommand{\arraystretch}{1.2}
  \begin{tabularx}{\linewidth}{l X c}
    \toprule
    \textbf{Metric} & \textbf{Definition} & \textbf{Illustrative goal} \\
    \midrule
    \textsc{aar}
      & (Agent-generated lines) $/$ (total merged lines)
      & ${>} 95\%$ (high-autonomy regimes) \\
    Human touchpoints
      & Manual edits after synthesis per merge window
      & $\rightarrow 0$ with spec corrections upstream \\
    Verification latency
      & Wall time for multi-agent audit / \textsc{ci} evidence to clear
      & Order of minutes (not days) \\
    \bottomrule
  \end{tabularx}
\end{table}

\subsection{Cognitive debt and mean time to repair}

Classically, mean time to repair (\textsc{mtt}) was bounded by how quickly a
human could navigate opaque code. Under SDAD, the binding limit is often how
fast accountable owners can revise the \textbf{specification} and re-run
synthesis plus verification. Useful proxies include:
\begin{itemize}[leftmargin=*,nosep]
  \item \textbf{Spec-to-product latency:} elapsed time from an approved business
    change request to a merged, released artefact faithful to the updated spec.
  \item \textbf{Cognitive gap index:} a qualitative score for ``shadow''
    technical debt---behaviourally correct agent-authored regions that no longer
    have human readers who could safely modify them without re-deriving intent
    from tests and specs, discussed further in
    Section~\ref{subsec:human-comprehension-risk}.
\end{itemize}

\begin{table}[htbp]
  \centering
  \small
  \caption{Era-level emphasis of dominant software metrics (illustrative).}
  \label{tab:metric-evolution}
  \renewcommand{\arraystretch}{1.2}
  \begin{tabularx}{\linewidth}{c l X X}
    \toprule
    \textbf{Era} & \textbf{Methodology} & \textbf{Primary metrics} & \textbf{Focus} \\
    \midrule
    1.0 & Waterfall & KLOC / function points & Volume of output \\
    2.0 & Agile/Scrum & Velocity / story points & Pace of iteration \\
    3.0 & SDAD & \textsc{ser}, spec fidelity, \textsc{aar} & Clarity of intent \\
    \bottomrule
  \end{tabularx}
\end{table}

\section{Quantitative Economics of Agentic Specification and Synthesis}
\label{sec:economics}

The shift toward SDAD changes not only \emph{how} teams build software but also
how capital is deployed. Synthesis latency contracts, while accountable effort
concentrates in specification, architecture, and verification. As a result,
variable implementation spend shifts from predominantly labour-driven execution
towards a mixed model in which inference cost and governance quality jointly
shape outcomes. Section~\ref{sec:sdad-metrics} introduced intent-validation
indicators; this section focuses on economic interpretation through indicative,
non-audited cost abstractions aligned with the enterprise scenario used in this
manuscript. \textit{Unless explicitly tied to cited empirical datasets, numeric
anchors are illustrative and intended to support reasoning rather than claim
universal rates.}

\subsection{Inference versus salary: marginal cost of code}

Under Human-Agile delivery, a useful abstraction for incremental implementation
spend is the \textbf{marginal cost of code} (\textsc{mcc}): the dollars required
to deliver the next unit of working, reviewed functionality under fixed team
structure and tooling. In this regime, the dominant variable is practitioner
time, and process controls are primarily tuned to labour capacity and flow.

Agentic pipelines reallocate part of that variable cost to billed
\textbf{inference}, where synthesis is priced through token consumption.
Empirical evidence indicates substantial declines in token pricing over recent
model generations, with heterogeneous behaviour across model tiers and
quality-adjusted settings~\cite{du2026,gundlach2025}. The economic implication
is not the disappearance of human cost, but a redistribution of scarce inputs:
from routine implementation effort towards specification quality,
verification discipline, and governance of iterative synthesis loops.

Productivity studies in industrial settings are directionally consistent with
this shift: AI-assisted development improves task throughput, while retaining
material oversight requirements~\cite{cui2025,paradis2024}. Accordingly, the
primary control objective moves from labour utilisation alone to intent
governance, with emphasis on reducing ambiguity and rework so that the
iterative multiplier \(\varphi\) in Equation~\eqref{eq:tci-agentic} remains
bounded. The following subsections quantify \textsc{tci}, \textsc{ser}, and
ledger-scale illustrations under this posture.

\subsection{From marginal human labour to total cost of implementation}

In a traditional Human-Agile baseline (circa\,2020--2024), the marginal cost of
adding implementation work was dominated by practitioner time and blended rate.
In the Agentic Era, we take the \textbf{total cost of implementation} (TCI) for a
given feature module to comprise both token-mediated synthesis loops and accountable
human architecture:

\begin{equation}
  \mathrm{TCI}_{\mathrm{agentic}}
  =
  \varphi\,
  \bigl(T_{\mathrm{in}} + T_{\mathrm{out}}\bigr)\,
  P_{\mathrm{token}}
  +
  \bigl(H_{\mathrm{arch}} \cdot R_{\mathrm{arch}}\bigr)
  \label{eq:tci-agentic}
\end{equation}

Here $\varphi$ is the iterative multiplier (\emph{e.g.\ }retries needed to reconcile
generated artefacts with an FRD), $T_{\mathrm{in}}$ and $T_{\mathrm{out}}$ are total
prompt and completion tokens (summed across loops if $\varphi$ counts whole passes),
and $P_{\mathrm{token}}$ is the effective marginal price per token for the chosen
tier and model class. For high-reasoning frontier routing in\,2026, an order-of-magnitude
anchor used in illustrative calculations below is roughly \textbf{USD\,0.05 per million}
mixed input/output tokens. The additive term $(H_{\mathrm{arch}}{\cdot}R_{\mathrm{arch}})$
represents billed Spec\,Architect oversight and gatekeeping retained in SDAD\@.

\subsection{Synthesis Efficiency Ratio (\textsc{ser})}

To connect specification quality to expenditure, define the \textbf{Synthesis
Efficiency Ratio} (\textsc{ser}):

\begin{equation}
  \mathrm{SER}
  =
  \frac{\text{\emph{Effective logic density of the specification}}}
       {\text{Total tokens expended}}
  \label{eq:ser}
\end{equation}

A high SER indicates that a large FRD is sufficiently deterministic relative to token
consumption (\emph{e.g.\ }minimal ambiguity rework), supporting what we might call a
\textbf{zero-waste synthesis} trajectory. Conversely, ambiguous requirements induce
\textbf{repair loops} (\emph{larger $\varphi$}, sometimes super-linear prompts) that burn
budget while still risking technical debt~\cite{he2026speed}. Coupled with
Equation~\eqref{eq:ambiguity_tax}, $\mathrm{SER}$ is consistent with treating crisp
requirements as simultaneously a quality knob and an economic control.

\subsection{Comparative scenario: enterprise ledger-scale module}

Table~\ref{tab:economics-erp} aggregates an illustrative balance sheet comparing a
capital-intensive Agile delivery trajectory (historical staffing pattern) versus an
SDAD-style path anchored on a deterministic FRD\@. Numeric examples are illustrative
order-of-magnitude anchors for exposition; calibrated enterprise studies should swap
organisation-specific blended rates and token telemetry.

\begin{table}[htbp]
  \centering
  {\small\caption{Illustrative TCI snapshot for a complex module sourced from a
    large (\(\sim\)1\,000\,page-class) specification ledger scenario.
    Figures are explanatory, not audited.
    Numeric entries reflect the author's own observational estimates and are not taken from cited sources.}}
  \label{tab:economics-erp}
  \renewcommand{\arraystretch}{1.22}
  \begin{tabularx}{\linewidth}{>{\raggedright\arraybackslash}X r r r}
    \toprule
    \textbf{Metric}
      & \textbf{Human-Agile (2020)}
      & \textbf{Agentic SDAD (2026)}
      & \textbf{Variance} \\
    \midrule
    Human labour
      & 480 person-hours
      & 4 architect-hours
      & \(\approx\)\,-99\% \\
    Calendar time (construction)
      & 4 weeks (\(\sim\)2 sprints)
      & 15\,min synthesis pass
      & \(\approx\)\,-99\% \\
    Delivery labour\newline\(\approx\) \$100/h blended
      & $\sim$\,USD\,48{,}000
      & $\sim$\,USD\,400
      & \(\approx\)\,-99\% \\
    Management overhead
      & $\sim$\,USD\,12{,}000
      & $\sim$\,USD\,600
      & \(\approx\)\,-95\% \\
    Token compute\newline(@\,USD\,0.05\,/\,10\textsuperscript{6} tok.; low $\varphi$)
      & ---
      & $\sim$\,USD\,0.20
      & --- \\
    \midrule
    \textbf{TCI}
      & \textbf{$\sim$\,USD\,60{,}000}
      & \textbf{$\sim$\,USD\,600}
      & \textbf{$\approx$\,-99\%} \\
    \bottomrule
  \end{tabularx}
\end{table}

\subsection{TCI curves: low \texorpdfstring{$\boldsymbol{\varphi}$}{phi} as economic necessity}

\begin{figure}[htbp]
  \centering
  \includegraphics[width=0.95\linewidth]{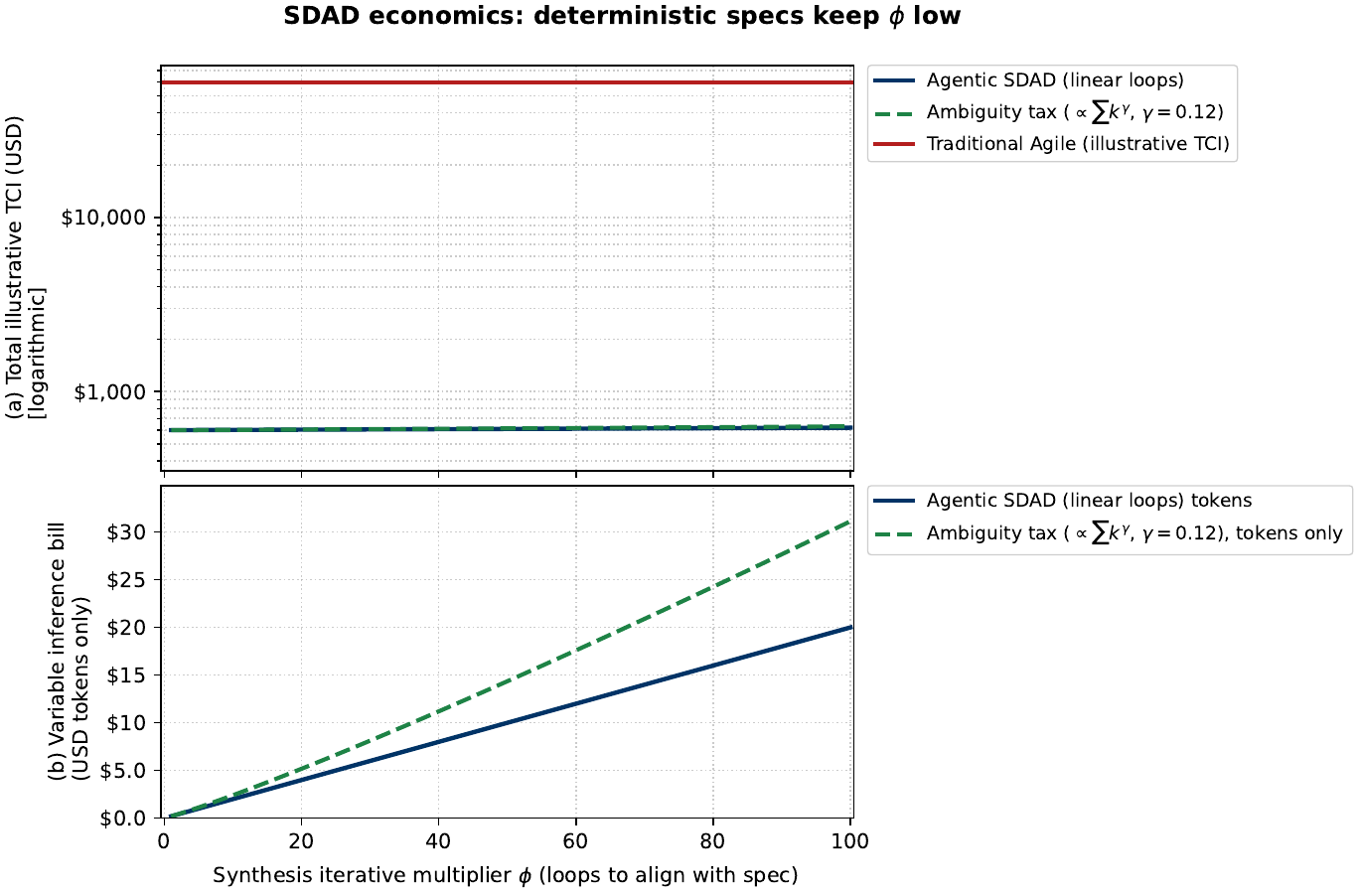}
  \caption{\textbf{(a)}~Illustrative total TCI (USD) versus multiplier
    $\varphi$, \emph{logarithmic} cost axis\,---\,agentic totals stay orders of
    magnitude below flat traditional Agile, while fixed oversight hides small
    token-driven slopes on this pane.
    \textbf{(b)}~Variable billed inference (\emph{i.e.}, token spend excluding
    the fixed oversight term): inference dollars rise with $\varphi$, with
    ambiguity ($\gamma{=}0{.}12$) bowing past linear loops.}
  \label{fig:economics-tci}
\end{figure}

Figure~\ref{fig:economics-tci} contrasts aggregate illustrative spend with token-only
variation as $\varphi$ grows; the takeaway is nonetheless explicit:
\textbf{returning synthesis work to deterministic, phased specifications is not mere
retro styling}---it aligns with shrinking $\varphi$ and stabilising SER
(Equation~\ref{eq:ser}). In budget terms, CIO offices therefore reallocate dollars from
\textbf{delivery headcount squads} toward \textbf{premium frontier access}, reliable
tiering for low-latency inference, observability/governance tooling, and a smaller cadre of
\textbf{logic engineers}: Spec Architects with both domain mastery and disciplined formal
articulation~\cite{otoum2026methods}.

\section{Hybrid Estimation in the SDAD Era}
\label{sec:hybrid-estimation}

In the era of Spec-Driven Agentic Development, the relationship between software
size, delivery time, and cost is no longer governed chiefly by human coding
labour. Estimation shifts upstream: the dominant variables become
\textbf{specification fidelity}, \textbf{review and governance effort}, and the
behaviour of the agentic synthesis loop under verification.

\subsection{Estimation Dynamics under SDAD}
\label{subsec:size-time-decoupling}

Historically, schedule growth tracked implementation labour. Under SDAD,
implementation is compressed into inference-time synthesis once the
specification is sufficiently precise; the critical path therefore moves from
\emph{construction} to \emph{intent resolution} and verification~\cite{royce1970}.
Accordingly, estimation should de-emphasise KLOC/function-point labour proxies
and prioritise specification quality, governance load, and expected loop
stability. The associated cost structure is similarly reweighted: architect and
assurance effort remain dominant fixed human terms, while synthesis becomes a
token-mediated variable term whose overruns are primarily ambiguity-driven
(Section~\ref{sec:economics}).

\subsection{Hybrid estimation as calibrated planning}
\label{subsec:hybrid-estimation-framework}

Hybrid estimation combines two complementary planning streams. First, it plans
human effort for requirements stabilisation, architectural decision-making,
governance, and release control. Second, it executes a bounded calibration pass
on representative work to measure throughput, token demand, and verification
behaviour under realistic conditions. This reframes Royce's ``do it twice''
principle for SDAD: the preliminary pass is not duplicate implementation, but an
evidence-generating rehearsal that surfaces ambiguity, unstable assumptions, and
weak SER prior to scale-up. Hybrid estimation is therefore a calibration
discipline, not a fixed process template; it preserves strategic planning while
anchoring forecasts in measured agentic performance.

\subsection{Comparative Estimation Metrics: Waterfall vs.\ Agile vs.\ Agentic}
\label{subsec:estimation-metrics-table}


\begin{table}[htbp]
  \centering
  \caption{Estimation metrics across SDLC paradigms. ``Spec Token Fidelity''
    denotes the proportion of specification statements that are unambiguous and
    directly machine-interpretable; ``Logic Density'' replaces story points as
    the primary complexity measure in SDAD.}
  \label{tab:estimation-metrics}
  \begin{tabularx}{\linewidth}{@{} l >{\raggedright\arraybackslash}X >{\raggedright\arraybackslash}X >{\raggedright\arraybackslash}X @{}}
    \toprule
    \textbf{Estimation Metric} & \textbf{Waterfall} & \textbf{Agile (2021)} & \textbf{Agentic SDLC (2026)} \\
    \midrule
    Primary variable   & KLOC / FP                  & Developer man-hours      & Spec token fidelity         \\
    Complexity measure & Function points            & Story points             & Logic density               \\
    Risk factor        & Requirement freeze failure & Human communication gap  & Model reasoning threshold   \\
    Cost basis         & Salary \& overhead         & Salary \& overhead       & Inference \& architect time \\
    Buffer management  & Phase contingency reserves & 20\,\% sprint buffer     & Multi-agent audit cycles    \\
    Estimation horizon & Months to years (upfront)  & Sprint-by-sprint (2 wks) & Pilot-calibrated             \\
    \bottomrule
  \end{tabularx}
\end{table}

Table~\ref{tab:estimation-metrics} summarises the categorical nature of this
shift. Waterfall and Agile estimate primarily around labour and coordination,
whereas SDAD estimates around specification quality, calibrated agentic execution,
and the human review structure that governs release.

\section{Discussion: Challenges and Ethics}
\label{sec:discussion}

\subsection{Maintaining human comprehension and operational readiness}
\label{subsec:human-comprehension-risk}

A repository synthesised by an Agentic Swarm may be formally correct, exercised
by generated tests, and aligned with verification reports, yet remain opaque to
a human engineer without the Specification as primary ground truth
(Section~\ref{sec:framework}). That gap creates ethical and operational risk:
on-call response under pressure, onboarding that must prioritise spec literacy
over informal code archaeology, and maintainability when design rationale is
concentrated in machine-readable contracts rather than in idiomatic human
authorship~\cite{hassan2026trustworthy}. It also interacts with the economics of
Section~\ref{sec:economics}: labour displaced from line-by-line coding reappears
as governance, incident ownership, and spec curation---categories that are easy
to under-budget if only token spend is tracked~\cite{cui2025,paradis2024}.

\citet{hassan2026trustworthy} argues that engineering organisations must invest
in ``Specification Archaeology'' tooling---AI-assisted reconstruction of intent
from legacy or agent-generated codebases---so that comprehension, audit, and
regulatory narrative keep pace with synthesis throughput.

\subsection{Security guardrails and provenance at synthesis scale}

The scale of autonomous multi-file generation widens attack surface and
amplifies specification mistakes. A malicious or carelessly ambiguous
specification can cause a synthesis agent to instantiate a single flawed pattern
across many services in minutes---for example one hallucinated authentication
shortcut propagated fleet-wide. SDAD mitigation aligns with
Section~\ref{sec:framework} (independent verification agents, sandboxed
execution, human sign-off) and with the testing literature's emphasis on not
delegating \emph{release authority} to the same model family that authored or
self-healed tests without redundancy~\cite{hassan2026trustworthy,pham2022}.

\begin{itemize}[noitemsep]
  \item \textbf{Specification-level security review:} Security and abuse
    requirements are first-class fields in the Formal Specification, not
    post-hoc tickets.
  \item \textbf{Independent security verification agents:} Adversarial review
    by agents separated from synthesis, mirroring the separation principle for
    critique in subsection~\ref{subsec:ai-testing}~\cite{hassan2026trustworthy}.
  \item \textbf{Sandboxed agentic execution:} Synthesis with minimal
    privilege, constrained egress, and auditable tool use.
  \item \textbf{Provenance tracking:} Cryptographic or policy-enforced linkage
    from merged artefacts to specification revision, model route, and gate
    outcomes---essential for forensics when $\varphi \gg 1$ repair loops or
    multi-agent handoffs obscure causality.
\end{itemize}

\subsection{Economic fairness, accountability, and vendor dependence}

Section~\ref{sec:economics} shows that token-priced synthesis can collapse
calendar construction time while $\mathrm{TCI}_{\mathrm{agentic}}$ remains
sensitive to ambiguity (large~$\varphi$), architect retainers, and premium
frontier routing~\cite{du2026,gundlach2025}. Ethically and organisationally,
three tensions deserve explicit treatment. \textbf{Labour dynamics:} Shifting
headcount from generalist implementers toward a smaller Spec Architect cadre
concentrates scarce, high-trust work; training pipelines and junior career paths
must be redesigned rather than assumed away~\cite{klieger2024chatcollab}.
\textbf{Accountability:} When failures arise from agentic output, regulators,
customers, and internal audit will ask \emph{which} human role owned the
specification, the gate policy, and the merge decision---SDAD makes those lines
clearer in principle but demands documented practice~\cite{hassan2026trustworthy}.
\textbf{Vendor and model dependence:} Organisations that route core synthesis
through closed APIs inherit pricing, availability, and policy shifts; ethical
governance therefore includes tiering strategy, exit plans, and reproducibility
of builds against pinned spec versions.

\subsection{Open questions}

Several issues remain under-specified in research and practice; they bridge
Section~\ref{sec:comparison} (especially subsection~\ref{subsec:communication-hierarchy}),
Sections~\ref{sec:economics}--\ref{sec:framework}, and the governance posture of
{SDAD} overall.

\begin{itemize}[noitemsep]
  \item \textbf{Quality and reliability of AI-generated code.} Evidence ties
    faster throughput to complexity and coverage regressions when specification
    discipline is weak~\cite{he2026speed}. Which combinations of property-based
    checks, formal attachments, cross-model redundancy, and human gate density
    are necessary \emph{and} proportionate at enterprise scale
    remains open~\cite{hassan2026trustworthy,otoum2026methods,wang2024}.
  \item \textbf{Measuring Spec Fidelity and SER.} Operational definitions and
    longitudinal baselines for Spec Fidelity scores and for $\mathrm{SER}$
    (Equation~\ref{eq:ser})---beyond illustrative scenarios---are needed so that
    teams can steer $\varphi$ and capital allocation with the same rigour once
    applied to story-point velocity~\cite{he2026speed}.
  \item \textbf{End-to-end autonomy versus human oversight.} Predominantly
    AI-authored pipelines without standing human involvement may reduce latency
    but raise risks of specification drift, correlated agent failures, and
    weakened accountability~\cite{salem2024survey,hassan2026trustworthy}.
  \item \textbf{Human--AI collaboration design.} Protocols for delegation,
    escalation, shared artefacts, and mixed initiative in agent swarms lag
    classical team norms~\cite{klieger2024chatcollab}; {SDAD}'s specs and gates
    are a start, not a complete sociology of work.
  \item \textbf{Closed-loop critique among agents.} Reviewer agents can catch
    defects early, yet oscillation, error reinforcement, and unclear stopping
    criteria complicate design: when does critique improve quality versus merely
    inflate token bills~\cite{otoum2026methods,hassan2026trustworthy}?
\end{itemize}

These questions motivate longitudinal adoption studies, calibrated enterprise
ledgers for $\mathrm{TCI}_{\mathrm{agentic}}$, and shared benchmarks for
multi-agent verification---themes aligned with open problems in the agentic
software-engineering literature~\cite{otoum2026methods}.

\subsection{Adoption pathway: a practitioner migration blueprint}
\label{subsec:migration-blueprint}

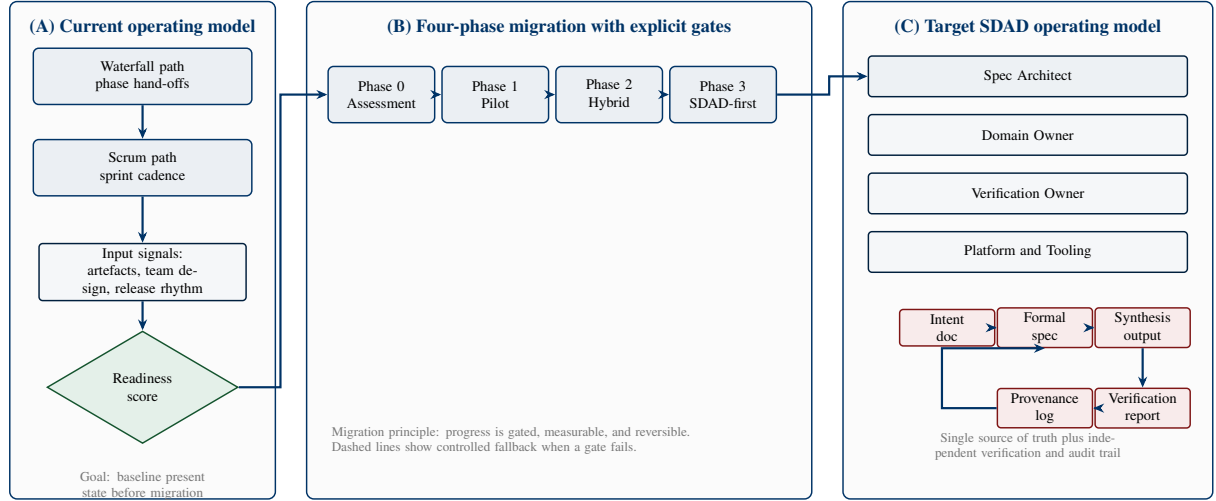
\begin{figure}[htbp]
  \centering
  \resizebox{\textwidth}{!}{
\begin{tikzpicture}[
  font=\scriptsize,
  >=Stealth,
  flow/.style={-{Stealth[length=6pt,width=5pt]}, very thick, darkblue},
  back/.style={-{Stealth[length=5pt,width=4pt]}, thick, dashed, medgray},
  panel/.style={draw=darkblue, thick, rounded corners=4pt, fill=lightgray!25, inner sep=7pt},
  phase/.style={draw=darkblue, thick, rounded corners=3pt, fill=darkblue!8, text width=3.45cm, minimum height=1.08cm, align=center, inner sep=3.5pt},
  gate/.style={diamond, aspect=1.7, draw=accentgreen!65!black, fill=accentgreen!14, thick, align=center, inner sep=0.9pt, text width=1.45cm},
  role/.style={draw=darkblue!60!black, thick, rounded corners=2pt, fill=white, text width=3.05cm, minimum height=0.78cm, align=center, inner sep=2.5pt},
  art/.style={draw=accentred!70!black, thick, rounded corners=2pt, fill=accentred!9, text width=2.75cm, minimum height=0.7cm, align=center, inner sep=2.5pt}
]

\node[panel, minimum width=5.1cm, minimum height=9.55cm, anchor=north west] (A) at (0,0) {};
\node[anchor=north, font=\footnotesize\bfseries, text=darkblue] at ($(A.north)+(0,-0.2)$)
  {(A) Current operating model};

\node[phase, text width=3.95cm] (wflow) at ($(A.north)+(0,-1.45)$)
  {Waterfall path\\phase hand-offs};
\node[phase, text width=3.95cm] (sflow) at ($(wflow.south)+(0,-1.2)$)
  {Scrum path\\sprint cadence};

\node[role, text width=3.8cm, fill=darkblue!4] (artin) at ($(sflow.south)+(0,-1.45)$)
  {Input signals:\\artefacts, team design, release rhythm};

\node[gate, text width=2.6cm, fill=accentgreen!12] (ready) at ($(artin.south)+(0,-1.65)$)
  {Readiness\\score};

\node[align=center, text width=4.2cm, medgray, font=\fontsize{7.4}{8.6}\selectfont]
  at ($(ready.south)+(0,-0.8)$)
  {Goal: baseline present state before migration};

\draw[flow] (wflow.south) -- (sflow.north);
\draw[flow] (sflow.south) -- (artin.north);
\draw[flow] (artin.south) -- (ready.north);

\node[panel, minimum width=9.7cm, minimum height=9.55cm, anchor=north west] (B) at (5.7,0) {};
\node[anchor=north, font=\footnotesize\bfseries, text=darkblue] at ($(B.north)+(0,-0.2)$)
  {(B) Four-phase migration with explicit gates};

\node[phase, text width=1.8cm] (p0) at ($(B.north west)+(1.45,-1.8)$) {Phase 0\\Assessment};
\node[phase, text width=1.8cm] (p1) at ($(p0.east)+(1.15,0)$) {Phase 1\\Pilot};
\node[phase, text width=1.8cm] (p2) at ($(p1.east)+(1.15,0)$) {Phase 2\\Hybrid};
\node[phase, text width=1.8cm] (p3) at ($(p2.east)+(1.15,0)$) {Phase 3\\SDAD-first};


\draw[flow] (p0.east) -- (p1.west);
\draw[flow] (p1.east) -- (p2.west);
\draw[flow] (p2.east) -- (p3.west);




\node[align=left, text width=8.7cm, medgray, font=\fontsize{7.2}{8.5}\selectfont, anchor=north west]
  at ($(B.south west)+(0.35,1.55)$)
  {Migration principle: progress is gated, measurable, and reversible.\\
   Dashed lines show controlled fallback when a gate fails.};

\node[panel, minimum width=7.1cm, minimum height=9.55cm, anchor=north west] (C) at (16.0,0) {};
\node[anchor=north, font=\footnotesize\bfseries, text=darkblue] at ($(C.north)+(0,-0.2)$)
  {(C) Target SDAD operating model};

\node[role, text width=5.95cm, fill=darkblue!7] (r1) at ($(C.north)+(0,-1.45)$) {Spec Architect};
\node[role, text width=5.95cm, fill=darkblue!4] (r2) at ($(r1.south)+(0,-0.72)$) {Domain Owner};
\node[role, text width=5.95cm, fill=darkblue!4] (r3) at ($(r2.south)+(0,-0.72)$) {Verification Owner};
\node[role, text width=5.95cm, fill=darkblue!4] (r4) at ($(r3.south)+(0,-0.72)$) {Platform and Tooling};

\node[art, text width=1.65cm] (a1) at ($(r4.south)+(-1.55,-1.05)$) {Intent\\doc};
\node[art, text width=1.65cm] (a2) at ($(a1.east)+(0.95,0)$) {Formal\\spec};
\node[art, text width=1.65cm] (a3) at ($(a2.east)+(0.95,0)$) {Synthesis\\output};
\node[art, text width=1.65cm] (a4) at ($(a3.south)+(0,-1.15)$) {Verification\\report};
\node[art, text width=1.65cm] (a5) at ($(a2.south)+(0,-1.15)$) {Provenance\\log};

\draw[flow] (a1.east) -- (a2.west);
\draw[flow] (a2.east) -- (a3.west);
\draw[flow] (a3.south) -- (a4.north);
\draw[flow] (a4.west) -- (a5.east);
\draw[flow] (a5.west) -- ++(-1.05,0) |- (a2.south);

\node[align=center, text width=6.2cm, medgray, font=\fontsize{7.2}{8.4}\selectfont]
  at ($(C.south)+(0,1.02)$)
  {Single source of truth plus independent verification and audit trail};

\draw[flow] (ready.east) -- ++(0.8,0) |- (p0.west);
\draw[flow] (p3.east) -- ++(1.0,0) |- (r1.west);

\end{tikzpicture}}
  \caption{Migration blueprint from Waterfall/Scrum to {SDAD} through gated
    phases, with controlled fallback loops and a target operating model centred
    on formal specification, independent verification, and provenance.
    {SDAD} adoption is staged and gate-driven, not a single-step replacement of
    Scrum or Waterfall.}
  \label{fig:sdad-migration-blueprint}
\end{figure}

The foregoing open questions are real, yet they should not paralyse teams that
wish to move now.
Figure~\ref{fig:sdad-migration-blueprint} distils a staged, gate-driven route
from Waterfall or Scrum into full {SDAD} practice into three panels.

Panel~A establishes the \emph{baseline}: both entry models contribute existing
artefacts, team design, and release rhythm to a structured readiness assessment.
The output is a readiness score that determines whether a team can proceed
to Phase~0 at all.

Panel~B sequences four migration phases separated by explicit, measurable
gates.
\textbf{Phase~0 (Assessment)} maps current toolchains and specification
habits against {SDAD} prerequisites.
\textbf{Phase~1 (Pilot)} runs a bounded workstream---one service or feature
track---under {SDAD} synthesis and gate discipline.
\textbf{Phase~2 (Hybrid)} widens scope while retaining legacy ceremonies for
teams not yet migrated, with a shared Verification Owner bridging both worlds.
\textbf{Phase~3 (SDAD-first)} retires sprint cadence for synthesised
deliverables and anchors daily work to Spec Fidelity and~$\mathrm{SER}$.
Between phases, four gates enforce measurable thresholds---Spec clarity,
Verification pass rate, Repair multiplier within band
(i.e.\ $\varphi \leq \varphi^{*}$), and Security gate pass---so that
progression is earned rather than scheduled.
Dashed feedback arrows allow controlled fallback when a gate fails.

Panel~C depicts the \emph{target operating model}: four role swimlanes
(Spec Architect, Domain Owner, Verification Owner, Platform and Tooling)
anchored by a five-artefact loop (Intent doc $\to$ Formal spec $\to$
Synthesis output $\to$ Verification report $\to$ Provenance log), with the
loop feeding back to the Formal spec to close the governance cycle.

\section{Conclusion}
\label{sec:conclusion}

The {SDLC} pendulum has not simply swung back to 1970s rigidity; long-context,
agentic tooling creates a new optimum in which \textbf{formal specification and
execution speed are complementary}. Humans still discover value through
iteration and feedback, but the implementation relay---once a chain of verbal
handoffs---compresses toward a machine-readable source of intent, explicit
verification gates, and orchestration policy (Section~\ref{subsec:communication-hierarchy}).
{Waterfall}-style discipline reappears as \textbf{BDUF}~2.0: not slow serial
document sign-off, but deterministic blueprints that keep repair
multiplier~$\varphi$ and technical-debt risk bounded while agents synthesise
code and tests at inference latency.

The {SDAD} framework presented here contributes both method and operating model:
four execution steps (Section~\ref{sec:framework}), a broader role
metamorphosis across engineering, QA, platform, and product functions
(subsection~\ref{subsec:team-metamorphosis}), intent-first metrics
(Section~\ref{sec:sdad-metrics}), quantitative economics
(Section~\ref{sec:economics}), hybrid estimation for planning
(Section~\ref{sec:hybrid-estimation}), and a staged migration blueprint for
adoption under measurable gates (subsection~\ref{subsec:migration-blueprint}).
Taken together, the argument is that organisations can scale agentic throughput
without surrendering governance if they make specification quality and
verification architecture first-class.

Core claims are:

\begin{enumerate}[noitemsep]
  \item \textbf{Ambiguity remains costly despite large contexts.} The Ambiguity
    Tax (Equation~\ref{eq:ambiguity_tax}) and low-$\mathrm{SER}$ repair loops
    make crisp, machine-readable requirements an economic control, not an
    optional ritual.
  \item \textbf{Engineering identity shifts upstream.} The software engineer as
    \emph{Spec Architect} owns domain fluency, formal articulation, adversarial
    specification, and gate policy---not keystroke throughput.
  \item \textbf{Velocity need not require quality collapse.} Empirical patterns link
    AI-accelerated delivery to rising debt when oversight is weak; {SDAD}-style
    specification and verification provide a path to throughput without the same
    regressions~\cite{he2026speed,otoum2026methods}.
  \item \textbf{Capital and metrics reorient.} Illustrative aggregates and
    Figure~\ref{fig:economics-tci} show how spend and attention move from
    staffing-heavy construction to retained architects, tiered inference, and
    tools that bound~$\varphi$; story points yield ground to {Spec Fidelity} and
    to $\mathrm{SER}$ as steering signals (Equation~\ref{eq:ser}).
  \item \textbf{Trust is architectural.} Independent verification agents,
    separation between synthesis and release authority, security-first specs, and
    human sign-off are integral to trustworthy agentic delivery---consistent
    with both the multi-agent gates in Section~\ref{sec:framework} and the
    testing and assurance literature
    surveyed in subsection~\ref{subsec:ai-testing}~\cite{hassan2026trustworthy,pham2022,wang2024}.
\end{enumerate}

Future work should operationalise {Spec Fidelity} and $\mathrm{SER}$ with
industry-calibrated measurement, benchmark hybrid-estimation accuracy across
domains, strengthen open toolchains for specification and provenance, and run
longitudinal studies of {SDAD} adoption spanning economics, security incidents,
and team structure.

\newpage
\bibliography{references}

\end{document}